\documentclass[sigconf]{IEEEtran}

\usepackage[utf8]{inputenc}
\usepackage[T1]{fontenc}
\usepackage{amsmath,amssymb,amsfonts}
\usepackage{graphicx}
\usepackage{subcaption} 
\usepackage{textcomp}
\usepackage{xcolor}
\newcommand{\mathbbm}[1]{\text{\usefont{U}{bbm}{m}{n}#1}} 
\usepackage{algorithm}
\usepackage{algorithmic}
\usepackage{multirow}
\usepackage{makecell}
\usepackage{mathtools}
\usepackage{enumitem}
\usepackage{diagbox}
\usepackage{subcaption}
\usepackage{array,colortbl}
\usepackage{dblfloatfix}
\usepackage{booktabs}

\newcommand*{\bt}{\boldsymbol{\theta}}

\title{Self-learn to Explain Siamese Networks Robustly}

\author
{
\IEEEauthorblockN{
Chao Chen\IEEEauthorrefmark{1},
Yifan Shen\IEEEauthorrefmark{2},
Guixiang Ma\IEEEauthorrefmark{3},
Xiangnan Kong\IEEEauthorrefmark{4},
Srinivas Rangarajan\IEEEauthorrefmark{5}, 
Xi Zhang\IEEEauthorrefmark{2}, 
Sihong Xie\IEEEauthorrefmark{1}
}
\IEEEauthorblockA{
\footnotesize
\IEEEauthorrefmark{1}Computer Science and Engineering Dept, Lehigh University 
\IEEEauthorrefmark{2}Laboratory of Trustworthy Distributed Computing and Service (MoE), BUPT,
\IEEEauthorrefmark{3}University of Illinois at Chicago,
\IEEEauthorrefmark{4}Worcester Polytechnic Institute,
\IEEEauthorrefmark{5}Department of Chemical and Biomolecular Engineering, Lehigh University
\\
chc517@lehigh.edu, 
shenyifan@bupt.edu.cn,
guixiang.ma@intel.com,
xkong@wpi.edu,
srr516@lehigh.edu,
zhangx@bupt.edu.cn,
xiesihong1@gmail.com
}
}

\begin{document}
\maketitle
\begingroup\renewcommand\thefootnote{\IEEEauthorrefmark{3}}
\footnotetext{This author now works at Intel Labs.}
\endgroup

\begin{abstract}
    Learning to compare two objects are essential in applications, 
    such as digital forensics, face recognition, and brain network analysis,
    especially when labeled data are scarce and imbalanced.
    As these applications make high-stake decisions and involve societal values like fairness 
    and transparency,
    it is critical to explain the learned models.
    We aim to study post-hoc explanations of Siamese networks (SN) widely used in learning to compare.
    We characterize the instability of gradient-based explanations due to the additional compared object in SN, 
    in contrast to architectures with a single input instance.
    We propose an optimization framework that derives global invariance from unlabeled data using self-learning to promote the stability of local explanations tailored for specific query-reference pairs.
    The optimization problems can be solved using gradient descent-ascent (GDA) for constrained optimization, or SGD for KL-divergence regularized unconstrained optimization, with convergence proofs, especially when
    the objective functions are nonconvex due to the Siamese architecture.
    Quantitative results and case studies on tabular and graph data from neuroscience and chemical engineering show that the framework respects the self-learned invariance while robustly optimizing the faithfulness and simplicity of the explanation.
    We further demonstrate the convergence of GDA experimentally.
\end{abstract}

\section{Introduction}
\label{sec:intro}

\begin{figure*}
    \centering
    \includegraphics[width=\textwidth]{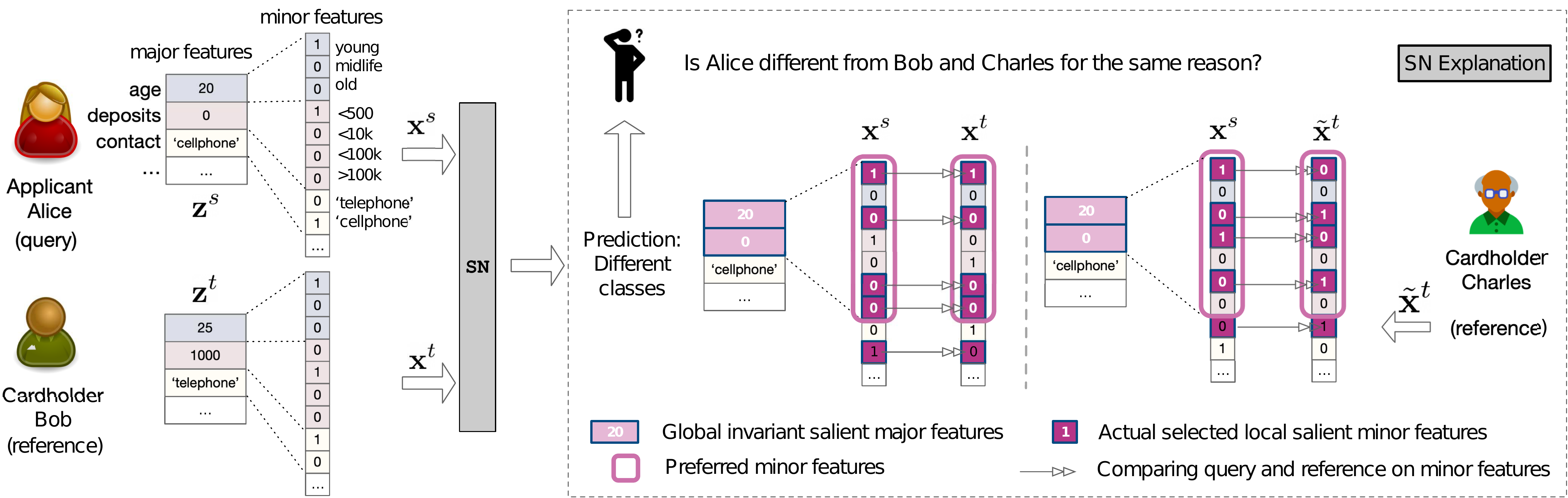}
    \caption{
    \small
    Explaining Siamese networks with invariant.
    \textit{Left:} major feature vectors $\mathbf{z}$ can be converted to minor feature vectors $\mathbf{x}$ by one-hot encoding,
    and the SN predicts the similarity of the query $\mathbf{x}^s$ and reference $\mathbf{x}^t$.
    \textit{Right:}
    SN explanation (SNX) for tabular data is enclosed in the dash-lined box.
    The workflow of SNX on graph data is similar but selects edges (see Fig.~\ref{fig:mask_demon}). 
    Global invariant salient features for Alice are \{\texttt{Age, Deposits}\},
    on both the major and minor levels (light purple values and boxes).
    The comparisons of Alice with references Bob and Charles lead to
    different local explanations (dark purple boxes), with two minor features selected beyond global invariant features.
    }
    \label{fig:overall}
\end{figure*}




Siamese networks (SN for short in the sequel)~\cite{chicco2021siamese} are widely used in similarity metric learning~\cite{Chopra2005,ma2019deep} and contrastive learning~\cite{chen20j,jaiswal2021survey} where objects are compared.
The applications of SN include digital forensics~\cite{Dey2017SigNetCS}, face recognition~\cite{Chopra2005}, 
and brain network analysis~\cite{ma2019deep}.
Different from conventional architectures that take one input instance,
an SN maps a pair of instances (the ``query'' and the ``reference'') to a similarity score~\cite{plummer2019these}.
As an example, in Fig.~\ref{fig:overall},
an SN is used to compare credit card applicants,
those of which differ
from the current cardholders will be rejected.
As SN is widely used in high-stake applications involving societal values,
it is urgent to provide simple and convincing explanations~\cite{goodman2017european},
to improve SN fairness and trustworthiness\cite{Lipton2018}.

\textbf{Challenges.}
We focus on post-hoc explanations of SN, represented by a small number of salient input elements that closely approximate the prediction made on the original input~\cite{ribeiro2016should,montavon2017explaining}.
On the one hand, in contrast to explaining architectures with one single input~\cite{ribeiro2016should,ying2019gnnexplainer},
explanations for SN should consider both query and reference,
and explanations insensitive to either one of them~\cite{plummer2019these} are misleading.
On the other hand, the additional reference can make the explanation over-sensitive to irrelevant perturbations.
A good explanation should be an invariant pertaining to the fixed query and reference.
For example, in Fig.~\ref{fig:overall},
with the same prediction (``\textit{different classes}'') on two query-reference pairs, 
Alice-Bob and Alice-Charles,
the corresponding explanations should differ as the reference changes from Bob to Charles.
However,
if the SN does encode certain invariant properties of Alice, 
both explanations should faithfully respect the invariance,
regardless of any \textit{superfluous} variations in the references.

Beyond tabular data in the illustrating example,
SN is used in graphs \cite{ma2019deep},
and thus the explanation for the query calls for invariant subgraphs, similarly.
Neuroscience studies have shown that the global Default Mode Network (DMN) \cite{raichle2015brain} consisting of several brain regions of interest (ROI) involves multiple cognitive and affective functions,
such as emotional processing and self-referential mental activity.
Researchers interested in the connections among the ROIs in a query bipolar patient~\cite{zovetti2020default} can require an invariant subgraph of the ROIs to be respected when explaining the difference between the query with healthy reference controls.
In molecules,
some specific chemical sub-structures lead to properties, 
such as solubility and insolubility \cite{Duvenaud2015},
and such sub-structures can be considered as the invariant of the molecule graphs.
However, 
in many scenarios, 
obtaining labeled data or domain knowledge to specify such invariants can be extremely costly.

\textbf{Proposed Method.}
To control superfluous variations,
we find invariants in the form of global salient features for each individual instance
using self-supervision on unlabeled data.
We then formulate a constrained optimization problem 
to adapt the invariant saliency map to explain an SN prediction local to a query-reference pair.
The adaptation balances the conformity to the invariance and the local flexibility when comparing a query to different references\footnote{
``Global'' means ``regardless of the references compared with a query'', rather than the universal behaviors of the explained model over the entire data space~\cite{luo2020parameterized}.}.
For example, in Fig.~\ref{fig:overall},
invariant features \{\texttt{Age,Deposits}\} characterizing Alice are refined to specific values as explanations local to different references.
Similarly, global invariant subgraphs of ROIs important to a subject
are adapted when comparing the subject with multiple references (see Fig.~\ref{fig:bp_case_study}).
We design a gradient descent ascent (GDA) algorithm SNX (SN Explainer)
to solve the constrained optimization problems.
Alternatively, we formulate an unconstrained optimization problem with KL-divergence regularization (SNX-KL)
to be solved by stochastic gradient descent (SGD).
The objective functions are nonconvex due to the SN architecture, 
and we prove the convergence of GDA based on 
nonconvex minimax optimization~\cite{lin20a}.

Based on the general framework,
the optimization problem can be specialized to incorporate additional constraints.
One-hot encoding is widely used on tabular datasets to help model optimization and
meet domain-specific requirements~\cite{kolesar1985robust}.
In Fig.~\ref{fig:overall}, we have three binary features (called ``minor'' features) to represent the three values (\textit{young}, \textit{midlife}, and \textit{old}) of the feature \texttt{Age} (called ``major'' features).
One-hot encoding places constraints over local explanations:
a major categorical feature is salient if and only if at least one of the associated minor binary features is salient.
Local explanations on the same query but different reference instances should select different binary minor features under the same major features (e.g., \texttt{Age} and \texttt{Deposits}) that globally characterize the query.
Prior methods explaining SN~\cite{plummer2019these,utkin2019explanation} are designed for images without such constraints.
See Section~\ref{sec:opt_tabular} for details.

Regarding graph data,
explaining the predicted similarity 
by subgraphs enumeration is NP-hard.
Recent graph explanation approaches treat the edges independently,
possibly leading to less coherent subgraphs
that are not interpretable,
as
larger connected subgraphs can have biological or chemical significance~\cite{ying2019gnnexplainer,jin2018junction}.
We introduce structural constraints to make adjacent edges more likely to be selected
into a subgraph as global invariant characterization of each graph, 
before finding local explanations 
between a query and reference graphs.
See Section~\ref{sec:opt_graph} for details.


\textbf{Contribution.}
1) We formulate the explanations of SN as two optimization problems.
2) We propose SNX-KL and SNX to solve the optimization problems with convergence guarantee theoretically and experimentally.
3) We demonstrate that self-supervised learning can find meaningful invariants to regulate local explanations,
and our algorithms outperform state-of-the-art explainers in six datasets with respect to faithfulness, counterfactual and conformity.
4) We analyze case studies on tabular and graph data.



\section{Problem Definition}

\subsection{Data with structures}
\label{sec:data_structure}
\textit{Tabular data} is a set of vectors, each with the same list of $q$ categorical \textit{major} features $\mathbf{z}=[z_1,\dots, z_q]$,
such as age and deposits of a credit card applicant~\cite{grath2018interpretable,kolesar1985robust}\footnote{Continuous features can be discretized into categorical features for the sake of explainable machine learning and domain-specific requirements.}.
Using one-hot encoding, each major feature $z_{i}$
is transformed to a set of binary \textit{minor} features
$x_{i,j} \in \{0,1\}$,
where $x_{i,j}=1$ if $z_i$ takes the $j$-th possible value.
As a result, $\sum_{j}x_{i,j}=1, \forall i=1,\dots,q$.
The \textit{minor} feature vector
$\mathbf{x} \in\{0,1\}^p$ is the concatenation of all binary minor features,
and $p$ is the number of minor features.
One-hot encoding allows mixed types of features, including special flags without ordinal semantics, to be treated uniformly, and explanations extracted from one-hot vectors are more actionable for recourse~\cite{Ustun2019},
for example, by telling an applicant to increase the deposit to ``$\geq500$'' rather than just ``the deposit amount causes the rejection''.
Continuous features in different scales can be discretized and one-hot-encoded to facilitate numerical optimization~\cite{bishop2006pattern}.


\textit{Graph data} is a set of graphs,
and each graph $G=(V,E)$
contains a set of vertices $V$ and edges $E\subset V\times V$.
We assume $G$ is undirected and its adjacency matrix $A$ is symmetric:
$A_{ij}=A_{ji}=1$ if 
nodes $v_i\in V$ and $v_j\in V$ are connected.
To unify the descriptions of optimization problems,
$A$ is flattened to a vector $\mathbf{x}$
of length 
$(|V|-1)(|V|-2)/2$
due to symmetry.
For different graphs, $\mathbf{x}$ can be of different lengths.
We adopt a GNN that considers node attributes,
but we focus on extracting subgraphs and retain all node features
and therefore do not explicitly denote node features.
Table \ref{tab:definitions} lists the symbols.

\subsection{Siamese Networks}
An SN accepts a pair of instances, denoted as query $\mathbf{x}^{s}\in\mathbb{R}^{p_s}$ and reference $\mathbf{x}^{t}\in\mathbb{R}^{p_t}$ \cite{plummer2019these},
with $p_s=p_t$ for two vectors, and $p_s\neq p_t$ in general for two different graphs.
The superscript $s$ or $t$ will be omitted when referring to a single instance in general.
The SN consists of a mapping function $emb(\mathbf{x}; \boldsymbol{\theta})$ that maps $\mathbf{x}^{s}$ and $\mathbf{x}^{t}$
to a latent space, where a metric measures the similarity between the two embeddings.
The mapping function $emb(\mathbf{x};\bt)$ can be an MLP for vectors and a GNN for graphs~\cite{Duvenaud2015}.
The similarity metric $sim(\cdot, \cdot)$ can be 
cosine similarity.
The SN is then the composite function
$f(\mathbf{x}^{s}, \mathbf{x}^{t}; \bt) = sim(emb(\mathbf{x}^{s};\bt), emb(\mathbf{x}^{t};\bt))$.
$f$ is trained to maximize (minimize, resp.) the similarity between any two instances of the same class (different classes, resp.) using some loss function $\ell^{\textnormal{SN}}$ as follows,
\begin{equation*}
\label{eq:sn}
\begin{aligned}
    \min_{\bt} & 
    \sum_{(s,t)\in\mathcal{T}} \ell^{\textnormal{SN}} (f(\mathbf{x}^{s}, \mathbf{x}^{t}; \bt), y_{st}), \\
\end{aligned}
\end{equation*}
where $\mathcal{T}$ is the training set containing all query-reference pairs.
The label of a pair $(\mathbf{x}^{s},\mathbf{x}^{t})$ is $y_{st}=\mathbbm{1}[y^{s} = y^{t}]$,
and is 1 if and only if the two instances have the same class label ($y^{s} = y^{t}$).

\begin{table}[t]
    \caption{\small Notation Definitions}
    \label{tab:definitions}
    \centering
    \small
    \begin{tabular}{|c||c|}
        \hline
        Notation & Definition\\\hline
        $ \mathbf{z}$ & A vector of $q$ categorical features in tabular data \\
        $G=(V,E)$ & A graph with sets of nodes $V$ and edges $E$ \\
        $A$ & Adjacent matrix of $G$\\
        $ \mathbf{x}$ & A vector of $p$ binary features encoding $\mathbf{z}$ or $A$ \\
        $f(\cdot,\cdot; \boldsymbol{\theta})$ & The target Siamese Network with parameter $\boldsymbol{\theta}$ \\
        $(\mathbf{x}^s,\mathbf{x}^t),y_{st}$ & Query and reference instances and the pair's label\\
        $\mathbf{N}, \mathbf{n}$ & Global and local masks over $\mathbf{z}$ (tabular data only)\\
        $\mathbf{M}, \mathbf{m}$ & Global and local masks over binary vector $\mathbf{x}$\\
        \hline
    \end{tabular}
\end{table}

\subsection{Post-hoc explanation of SN}
\label{sec:xml}

\begin{figure}[t]
    \centering
    \includegraphics[width=0.45\textwidth]{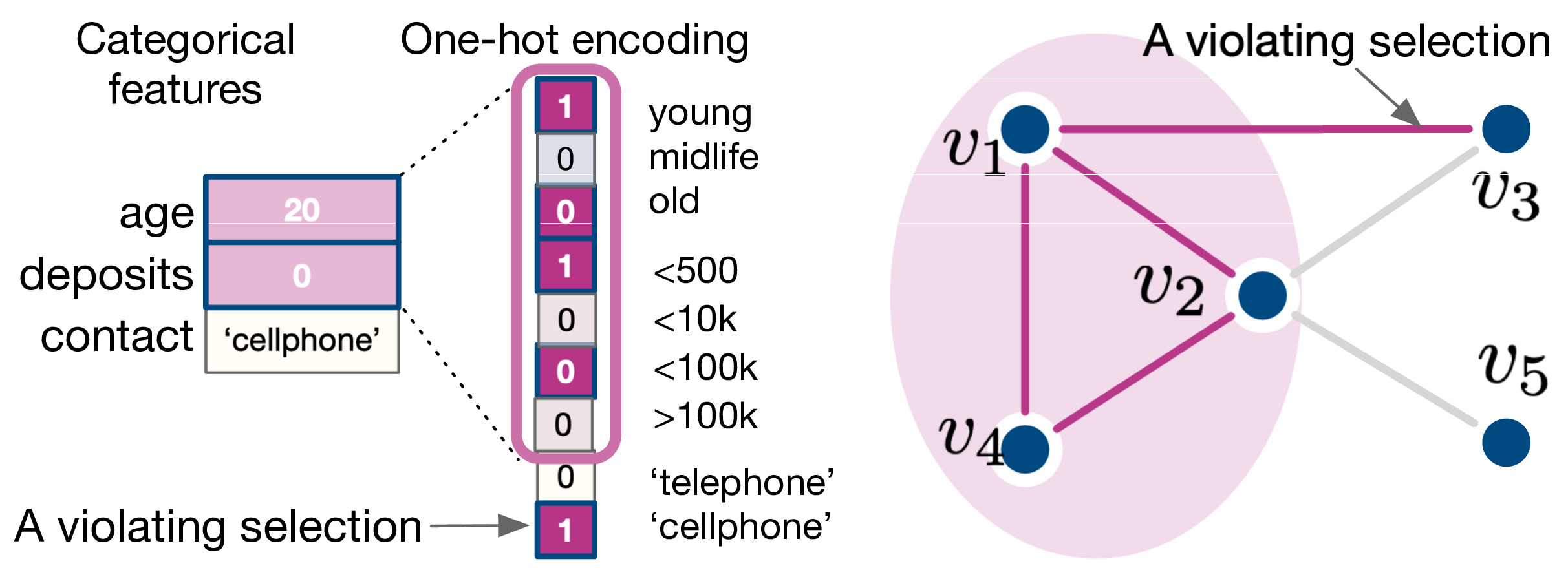}
    \caption{
    \small
        \textit{Left:} a vector of major features, with globally salient features in light purple and the corresponding minor feature vector,
        with local salient features in dark purple.
        \textit{Right:} the clique in light purple is a globally salient subgraph, while an additional edge $(v_1,v_3)$ is considered salient when a reference graph is compared.
    }
    \label{fig:mask_demon}
\end{figure}

We assume a trained SN $f(\mathbf{x}^{s}, \mathbf{x}^{t}; \bt)$ and focus on explaining the SN's predictions on test data.
The parameter $\bt$ is fixed and thus omitted from $f(\mathbf{x}^{s}, \mathbf{x}^{t}; \bt)$ when there is no confusion.
Given a pair of query
$\mathbf{x}^s \in\{0,1\}^{p_s}$ and reference
$\mathbf{x}^t \in \{0,1\}^{p_t}$,
$\mathbf{m}^s \in [0,1]^{p_s}$ and
$\mathbf{m}^t \in [0,1]^{p_t}$ are the corresponding multiplicative masks.
A large element in a mask indicates that the corresponding feature value contributes more to the SN prediction~\cite{ying2019gnnexplainer, selvaraju2017grad}.
The element-wise product $\mathbf{m} \otimes \mathbf{x}$ is a masked instance so that $\mathbf{m}_i \mathbf{x}_i\in[0,1]$ is the importance/saliency of the $i$-th element of $\mathbf{x}$.
Fig.~\ref{fig:mask_demon} demonstrates masked vector and graph instances.
Additive perturbations~\cite{utkin2019explanation, ribeiro2016should, lundberg2017unified} are less interpretable, as
the binary features can perturbed to outside the range $[0,1]$.
A prediction $f(\mathbf{x}^{s}, \mathbf{x}^{t})$ depends on both inputs $(\mathbf{x}^{s},\mathbf{x}^{t})$, so does 
the prediction's explanation.
Such dependencies lead to robustness issues of the gradient-based explanations.

\vspace{.05in}
\noindent\textbf{Robustness of SN explanations.}
Using a simple example SN $f(\mathbf{x}^{s}, \mathbf{x}^{t}; \bt)=\sigma(<\bt^\top \mathbf{x}^{s}, \bt^\top \mathbf{x}^{t}>)$, we characterize the robustness of gradient-based explanations of SN.
Taking the gradient of $\ell^{\textnormal{SN}}
$ with respect to the query $\mathbf{x}^{s}$,
we obtain a saliency map over $\mathbf{x}^{s}$ proportional to $\bt\bt^\top \mathbf{x}^t$.
The salient map explains the prediction using the magnitudes of elements in $\bt\bt^\top \mathbf{x}^t$ and depends on the SN parameter $\bt$ and the reference $\mathbf{x}^{t}$.
The saliency map can be manipulated to any pre-defined target explanation $\tilde{\mathbf{m}}^s$,
by perturbing the reference $\mathbf{x}^t$:
\begin{equation*}
\label{eq:snx_manipulation}
    \textnormal{min}_{\boldsymbol{\delta}\in\mathbf{R}^{p_t}}
    \|(\bt\bt^\top)(\mathbf{x}^t + \boldsymbol{\delta})-\tilde{\mathbf{m}}^s\|_2^2\hspace{.2in}
    \textnormal{s.t.}
    <\mathbf{x}^s, (\bt\bt^\top )(\boldsymbol{\delta})> = 0.
\end{equation*}
The objective pushes the saliency map to the target mask $\tilde{\mathbf{m}}^s$~\cite{Dombrowski2019,Ghorbani2017},
while the equality constraint specifies the orthogonality and that the SN prediction is not changed.
Simple algebra manipulations lead to the problem of finding a vector $\mathbf{x}^\perp$ that is orthogonal to $\mathbf{x}^s$.
Since
the one-hot encoding or a large number of disconnected pairs of vertices on a graph results in a large number of zeros in $\mathbf{x}^s$,
there are many vectors orthogonal to $\mathbf{x}^s$.
$\boldsymbol{\delta}$ can then be found by minimizing the loss $\|(\bt\bt^\top)\boldsymbol{\delta} - \mathbf{x}^\perp\|_2^2$ without constraint.

\vspace{.05in}
\noindent\textbf{Desiderata}. We aim to find local explanations (i.e., masks, 
denoted by $\mathbf{m}$ in general),
with the following merits: 
\begin{itemize}[leftmargin=*]
\item 
\textit{Simplicity}~\cite{ribeiro2016should} of an explanation derived from $\mathbf{m}$ is measured by the number of features or edges of most important according to $\mathbf{m}$,
or $\|\mathbf{m}\|$, the $\ell_1$ norm of $\mathbf{m}$.
\item
\textit{Faithfulness}~\cite{ribeiro2016should, lundberg2017unified}
can be evaluated by feeding the masked instances $(\mathbf{m}^s \otimes \mathbf{x}^s, \mathbf{m}^t \otimes \mathbf{x}^t)$ to the target SN $f$ and measuring the distortion in the original output:
\begin{equation}
\label{eq:faith}
    \ell(f(\mathbf{x}^s, \mathbf{x}^t), 
    f(\mathbf{m}^s \otimes \mathbf{x}^s, 
    \mathbf{m}^t \otimes \mathbf{x}^t)),
\end{equation}
where $\ell$ is some loss functions, 
such as the binary cross-entropy.
A low faithfulness loss indicates that the masks can select salient features to preserve the SN output $f(\mathbf{x}^s, \mathbf{x}^t)$.
Faithfulness and simplicity are conflicting goals:
the all-one mask retains all salient features but will lose simplicity.
\item
\textit{Counterfactuals} (CF) can be more interpretable and helpful for algorithmic recourse 
\cite{Wachter2017}.
A CF explanation is the complement $(1-\mathbf{m})$ of a given explanation $\mathbf{m}$ and can 
show ``what'' would the predictions be ``if'' 
keeping the non-salient features.
In Fig.~\ref{fig:mask_demon}, the counterfactual explanations are the non-purple elements.
We define the following \textit{counterfactual loss} to measure how much $\mathbf{m}^s$ and $\mathbf{m}^t$ fail to preserve:
\begin{equation}
\label{eq:cf}
    \ell(f(\mathbf{x}^s, \mathbf{x}^t), 
    f( (1-\mathbf{m}^s) \otimes \mathbf{x}^s, 
    (1-\mathbf{m}^t) \otimes \mathbf{x}^t)).
\end{equation}
If the masks have high weights over \textit{all} salient features, then their complements should distort the original SN prediction $f(\mathbf{x}^s, \mathbf{x}^t)$ most, with a high counterfactual loss.
A faithful mask may contain some but not all salient features, thus can preserve the SN prediction $f(\mathbf{x}^s, \mathbf{x}^t)$ while missing some redundant salient features with a low counterfactual loss.

\item
\textit{Conformity} measures how much a local explanation 
overlaps the global salient features.
Fig.~\ref{fig:mask_demon} shows that an additional minor feature or edge can be selected beyond the global invariant explanations to accommodate local query-reference pairs.
Conforming to an invariant leads to more robustness against variations in the reference $\mathbf{x}^t$. 
\end{itemize}
\section{Self-learn to explain robustly}
\label{sec:methodology}
We propose optimization problems and algorithms to generate explanations robustly without supervision. 


\subsection{A general optimization formulation}
\label{sec:general_opt}
The variables to be optimized are the two masks $\mathbf{m}^s \in [0,1]^{p_s}$ and $\mathbf{m}^t \in [0,1]^{p_t}$ over query $\mathbf{x}^s$ and reference $\mathbf{x}^t$, respectively\footnote{The sigmoid function is a monotonic bijective function with the range $[0,1]$,
we let the masks be the output of the sigmoid function.
}.
To balance the faithfulness and simplicity,
we have the objective:
\begin{equation}
\begin{aligned}
    \min_{\mathbf{m}^s, \mathbf{m}^t}  
    \ell \left( 
    f(\mathbf{x}^s, \mathbf{x}^t),
    f(\mathbf{m}^s \otimes \mathbf{x}^s, 
    \mathbf{m}^t \otimes \mathbf{x}^t) \right) \\
    +\gamma(\| a(\mathbf{m}^s) \| + \| a(\mathbf{m}^t) \|),
    \label{eq:obj_lm}
\end{aligned}
\end{equation}
where
$\gamma $ is a hyperparameter to fine-tune the relative importance of the two goals.
$\ell(\cdot,\cdot)$ as defined by Eq. (\ref{eq:faith}) promotes faithfulness
and $\|\cdot\|$ is the $\ell_1$-norm that promotes simplicity.
Note that simplicity can be structural (such as joint sparsity~\cite{Simon13asparse-group,friedman_sparse_2008}) and
the auxiliary function $a(\mathbf{m})$ maps from an unstructured mask to another vector,
upon which structural sparsity constraints
can be imposed on the tabular data (Section \ref{sec:opt_tabular}) and graph data (Section \ref{sec:opt_graph}).
%

\vspace{.1in}
\noindent\textbf{Stage 1}.
Saliency maps may lack robustness and
we use a global invariant to regulate $\mathbf{m}^s$ for more robustness against varying 
$\mathbf{x}^t$.
If there is domain knowledge regarding which features/edges in a query $\mathbf{x}$ are salient,
we can set the binary values in the global mask $\mathbf{M}\in[0,1]^p$ for $\mathbf{x}$ accordingly.
There is no such knowledge in a more general case, 
and we propose to extract $\mathbf{M}$ 
as an invariant to encode global salient elements in $\mathbf{x}$, regardless of references,
using
self-supervision learning (SSL).
SSL \cite{you2020graph,jin2020self} train a predictive model $h$ by contrasting $\mathbf{x}$ and its transformation 
$T(\mathbf{x})$, 
where $\mathbf{x}$ can be a graph \cite{you2020graph,jin2020self} or an image~\cite{chen20j}.
The objective function in SSL is:
\begin{equation*}
h^\ast 
= \min_h \mathcal{L}_{self} (T, \mathbf{x}, h) 
= \min_h \ell (h(\mathbf{x}), h(T(\mathbf{x}) ) ),
\end{equation*}
where $T(\cdot)$ is a given transformation.
For example, 
$T(\cdot)$ can be
cropping and rotating of images~\cite{chen20j}.
The most relevant transformation to our work is to use random walk to mask out irrelevant parts of a graph~\cite{jin2020self}, 
and $T(\mathbf{x})=\mathbf{M}\otimes \mathbf{x}$.
SSL uses a fixed $T$ function to learn $h$, while we are interested in learning a $T$, 
which is a global mask for $\mathbf{x}$ regardless of different references compared using a fixed SN model $f$.
We formulate the following optimization problem:
\begin{eqnarray}
    \min_{\mathbf{M}}
    & \ell \left(
    f(\mathbf{x}, \mathbf{x}),
    f(\mathbf{x}, \mathbf{M} \otimes \mathbf{x})
    \right) 
    + \gamma \|a(\mathbf{M})\|,
    \label{eq:obj_gm}
    \\
    & \textnormal{s.t.} \quad g_i(\mathbf{M})\leq0, i =1,\dots,c.
    \label{eq:obj_gm_con}
\end{eqnarray}
$\mathbf{M}$ is expected to extract the salient features of $\mathbf{x}$ to maximally preserve information in $\mathbf{x}$, indicated by a low faithfulness loss (Eq. (\ref{eq:faith})) when comparing $\mathbf{x}$ and $\mathbf{M} \otimes \mathbf{x}$.
The constraint functions $g_i(\mathbf{M})$ will be specified for tabular and graph data in the following two sections. 
Unsupervised feature selection~\cite{wei2017rethinking} learns important features without considering the SN architecture
and has inferior performance in the experiments. 

\vspace{.1in}
\noindent\textbf{Stage 2}.
After finding $\mathbf{M}^s$ and $\mathbf{M}^t$ for $\mathbf{x}^s$ and $\mathbf{x}^t$, 
respectively, using Eqs. (\ref{eq:obj_gm})-(\ref{eq:obj_gm_con}), we fix the global masks $\mathbf{M}^s$ and $\mathbf{M}^t$  and incorporate them as constraints in the following optimization problem to find local masks $\mathbf{m}^s$ and $\mathbf{m}^t$:
\begin{equation}
\begin{aligned}
    \min_{\mathbf{m}^s, \mathbf{m}^t} 
    \ell \left( 
    f(\mathbf{x}^s, \mathbf{x}^t),
    f(\mathbf{m}^s \otimes \mathbf{x}^s, 
    \mathbf{m}^t \otimes \mathbf{x}^t)
    \right) \\
    + \gamma \left(\|a(\mathbf{m}^s)\| + \|a(\mathbf{m}^t)\| \right)
    \label{eq:con_lm_obj}
\end{aligned}
\end{equation}
\begin{align}
    \textnormal{s.t.} 
    & \quad
    g_i(\mathbf{m}) = a(\mathbf{m}^s)_i - a(\mathbf{M}^s)_i \leq 0, 
    \: i =1, \dots, c_s,
    \label{eq:con_lm_con_1}\\
    & \quad 
    g_{c_s+i}(\mathbf{m}) = a(\mathbf{m}^t)_i - a(\mathbf{M}^t)_i \leq 0, 
    \: i =1, \dots,  c_t
    \label{eq:con_lm_con_2}.
\end{align}
$a(\cdot)_i$ means the $i$-th element of $a(\cdot)$,
and $c_s$ and $c_t$ are the number of the constraints derived from the auxiliary function due to structural sparsity.
Notice that $a(\cdot)$ is monotonic,
so that structural sparsity in the global mask $\mathbf{M}$ enforces structural sparsity in the local masks $\mathbf{m}$.

\subsection{Optimization problem for tabular data}
\label{sec:opt_tabular}
\noindent \textbf{Stage 1.}
Without a particular reference $\mathbf{x}^t$,
a global mask over a query can at best identify salient major features, such as \texttt{Age}, in  $\mathbf{z}^s$.
We use the auxiliary function to find a global invariant mask $\mathbf{N}
\in[0,1]^{q}$ over the \textit{major} features,
where $\mathbf{N}_{i}=a(\mathbf{M})_i=1 - \prod_j (1-\mathbf{M}_{i,j})$ is the importance of the $i$-th major feature\footnote{We tried alternatives, such as $\mathbf{N}_i = \sum_k \mathbf{M}_{i,k}$
and $\mathbf{N}_i =\prod_k \mathbf{M}_{i,k}$.
They cannot focus on minor features for a significant major feature, or lead to numerical underflow issues.},
and $\mathbf{M}_{i,j}$ indicates the global importance of the $j$-th value of the $i$-th major categorical feature of the query $\mathbf{x}$.
As we already encode the dependencies among minor features 
in $a(\mathbf{M})$,
there is no more constraints in Eq. (\ref{eq:obj_gm_con}) ($c=0$).
\noindent \textbf{Stage 2.}
Comparing with $\mathbf{x}^t$,
we further identify salient minor features, 
such ``\texttt{Age}<25'', 
associated with the salient major features.
For tabular data, any two input vectors to SN are aligned,
so we optimize a single mask $\mathbf{m}=\mathbf{m}^s=\mathbf{m}^t$ to find salient features for both instances.
We use the same auxiliary function for the local masks $\mathbf{n} = a(\mathbf{m})$ such that $\mathbf{n}_i = 1 - \prod_j (1-\mathbf{m}_{i,j})$
in Eq. (\ref{eq:con_lm_obj}).
As we focus on finding masks for the query $\mathbf{x}^s$ with varying references $\mathbf{x}^t$, only Eq. (\ref{eq:con_lm_con_1}) is kept ($c_s=q$, the number of major features).
Alternatively, 
we formulate an unconstrained optimization problem:
\begin{equation}
\label{eq:kl_reg}
 \min_{\mathbf{m}} \ell \left(
    f(\mathbf{x}^s, \mathbf{x}^t),
    f(\mathbf{m} \otimes \mathbf{x}^s, \mathbf{m} \otimes \mathbf{x}^t)
    \right) 
    + \gamma \|\mathbf{n} \| 
    + \beta \textnormal{KL}(\mathbf{n} \| \mathbf{N}),
\end{equation}
where 
$\textnormal{KL}(\mathbf{n} \|  \mathbf{N})=\sum_{i=1}^q \textnormal{KL}(\mathbf{n}_i\| \mathbf{N}_i)$ and 
$\textnormal{KL}(\mathbf{n}_i\| \mathbf{N}_i)$ is the KL-divergence between $\mathbf{n}_i$ and $\mathbf{N}_i$, treated as the means of two binary random variables. 
According to~\cite{bishop2006pattern} (Section 10.1),
the KL-regularization encourages $\mathbf{n}_i$ to be smaller than $\mathbf{N}_i$.

\subsection{Optimization problems for graphs}
\label{sec:opt_graph}
We set $a(\mathbf{M})=\mathbf{M}$ for masks on graphs.
Isolated single-edged subgraphs are not only difficult for domain experts to interpret, 
but can also disturb the working of the GNN within the SN.
Therefore, the selection of two adjacent edges should be related.
We consider such dependencies in stage 1, 
where the constraints in Eq. (\ref{eq:obj_gm_con}) encourage the connectivities,
\begin{align}
    g_{jk}(\mathbf{M}) = \| \mathbf{M}_j - \mathbf{M}_k \| - \epsilon \leq 0,j,k\textnormal{ adjacent in }G, \label{eq:con_gm_graph_connect}
\end{align}
where $\mathbf{M}_j\in[0,1]$ is the mask for the $j$-th edge.
The constraints indicate that the selection of the $j$-th edge can lead to the selection of the $k$-th edge if they share a node~\cite{luo2020parameterized},
and $\epsilon$ controls the co-occurrence of the two edges.
After obtaining $\mathbf{M}^s$ and $\mathbf{M}^t$ for each graph,
the local masks $\mathbf{m}^s$ and $\mathbf{m}^t$ over $\mathbf{G}^s$ and $\mathbf{G}^t$ are optimized by solving problem Eqs. (\ref{eq:con_lm_obj})-(\ref{eq:con_lm_con_2}),
using $\mathbf{M}^s$ and $\mathbf{M}^t$ as constants in the constraints.
In general, $\mathbf{G}^s$ and $\mathbf{G}^t$ have different numbers of nodes, 
thus $\mathbf{m}^s$ and $\mathbf{m}^t$ can lead to different numbers of constraints 
 ($c_s=p^s$ and $c_t=p^t$).
Similar to Eq. (\ref{eq:kl_reg}), 
one can use KL-regularization terms to enforce the constraints. 

\subsection{Optimization algorithm and convergence}
If we adopt the KL-regularization to incorporate global invariance as in Eq. (\ref{eq:kl_reg}) (referred to as SNX-KL),
SGD can be used and
the global masks act as regularization rather than hard constraints.
Since the objectives are not convex due to SN architecture, only convergence to local optima can be characterized, as in non-convex optimization~\cite{Ma2020WhyDL}. 

Alternatively,
global explanations on graph data (Eq. (\ref{eq:obj_gm})-(\ref{eq:obj_gm_con}))
and local explanations on both tabular and graph data (Eq. (\ref{eq:con_lm_obj})-(\ref{eq:con_lm_con_2}))
can be found by solving constrained optimization problems.
We adopt the gradient descent-ascent (GDA) algorithm~\cite{lin20a} to allow violations of the constraints.
Take optimizing local masks as an example,
the mask to optimize is
$\mathbf{m}=\mathbf{m}^s=\mathbf{m}^t$ for tabular data
and $\mathbf{m} = [\mathbf{m}^s; \mathbf{m^t}]$ (the concatenation of $\mathbf{m}^s$ and  $\mathbf{m^t}$) for graph data.
The objective function $g_0(\mathbf{m})$ is that in Eq. (\ref{eq:obj_gm}) or Eq. (\ref{eq:con_lm_obj}),
and the inequality constraints $g_i(\mathbf{m}), \forall i \in \{1, \dots, c \}$ are those defined in or Eq. (\ref{eq:obj_gm_con}) or Eqs. (\ref{eq:con_lm_con_1})-(\ref{eq:con_lm_con_2}),
with $c$ being the total number of inequality constraints in each problem.
In Eqs. (\ref{eq:con_lm_con_1}) - (\ref{eq:con_lm_con_2}), 
$c=q$ in tabular data,
and $c=p^s+p^t$ in graph data.
We introduce the non-negative Lagrange multipliers $\boldsymbol{\lambda}  \in\mathbb{R}_+^{c}$
and construct the Lagrangian
\begin{equation}
\label{eq:lagrangian}
\mathcal{L}(\mathbf{m}, \boldsymbol{\lambda}) 
= g_0(\mathbf{m}) + \sum_{i=1}^{c} \lambda_i g_i(\mathbf{m}),
\end{equation}
Then gradient descent is applied to $\mathbf{m}$ and gradient ascent is applied to $\boldsymbol{\lambda}$ with learning rates $\eta_1$ and $\eta_2$:
\begin{equation}
\label{eq:variable_update}
    \mathbf{m} \leftarrow \mathbf{m} - \eta_1  
    \frac{\partial \mathcal{L}}{\partial \mathbf{m}},
    \hspace{.2in}
    \boldsymbol{\lambda} \leftarrow \boldsymbol{\lambda} + \eta_2 
    \frac{\partial \mathcal{L}}{\partial \mathbf{\boldsymbol{\lambda}}}.
\end{equation}
Between the two updates, we use the latest $\mathbf{m}$ to evaluate the partial derivatives with respect to  $\boldsymbol{\lambda}$.
Also, the $\boldsymbol{\lambda}$ vector is normalized to have length one before entering the next update iteration.
The GDA-based SN explanation (SNX) algorithm is given in Algorithm~\ref{alg:con_opt}.
The time complexity of each optimization iteration is the sum of that of training the SN using back-propagation and that of evaluating the $c$ constraints.

\begin{algorithm}[t]
\small
\caption{SNX: Siamese Network Explanation with GDA}
\label{alg:con_opt}
\begin{algorithmic}[1]
\STATE \textbf{Input}: 
a target SN model $f$, 
a query instance $\mathbf{x}^s$ and reference instance $\mathbf{x}^t$,
\textit{optional} human-defined constraints in $\mathbf{M}^s$ and $\mathbf{M}^t$,
learning rate $\eta_1, \eta_2$ for $\mathbf{m}$ and $\boldsymbol{\lambda}$.
\STATE \textbf{Output}:
local masks $\mathbf{m}^s$ for $\mathbf{x}^s$, and $\mathbf{m}^t$ for $\mathbf{x}^t$.
\STATE \textbf{Init}:
$\boldsymbol{\lambda} = [1/c,\dots,1/c]\in\mathbb{R}^{c}$.
\IF{$\mathbf{M}^s$ and $\mathbf{M}^t$ not given}
\STATE Extract global masks $\mathbf{M}^s$ and $\mathbf{M}^t$ by 
Eqs. (\ref{eq:obj_gm})-(\ref{eq:obj_gm_con})
\hfill{$\rhd$SSL}
\ENDIF
\STATE Pretrain 
$\mathbf{m}^{s}, \mathbf{m}^t$ without constraints 
using Eq. (\ref{eq:con_lm_obj}).
\STATE Solve the full constrained optimization problems 
Eqs. (\ref{eq:con_lm_obj})-(\ref{eq:con_lm_con_2})
using GDA to find local masks $\mathbf{m}^s$ and $\mathbf{m}^t$.
\end{algorithmic}
\end{algorithm}

\noindent\textbf{Convergence}.
The Lagrangian is nonconvex-concave, in fact, linear in the dual variables.
The GDA algorithm is convergent based on Theorem 4.4 of the work~\cite{lin20a}, given the Lagrangian satisfies their Assumption 4.2, reproduced below:

\begin{enumerate}[leftmargin=*]
    \item 
\textit{ $\mathcal{L}$ is $l$-smooth and $\mathcal{L}(\cdot,\boldsymbol{\lambda})$ is $L$-Lipschitz for each $\boldsymbol{\lambda}$ and $\mathcal{L}(\mathbf{m},\cdot)$ is concave for each $\mathbf{m}$.
}
    \item
\textit{The domain of $\boldsymbol{\lambda}$ is convex and bounded.}
\end{enumerate}

We verify these assumptions:
if the target SN is a composition of smooth function (excluding functions such as ReLU), then $\mathcal{L}$ is smooth in both the primal variable $\mathbf{m}$ and the dual variable $\boldsymbol{\lambda}$;
since the masks are all non-negative, the $\ell 1$ penalties turn out to be just the sum of the elements in the masks.
Due to the smoothness, $\mathcal{L}(\cdot,\boldsymbol{\lambda})$ is $L$-Lipschitz. 
$\mathcal{L}$ is linear and thus concave in $\boldsymbol{\lambda}$.
In Algorithm~\ref{alg:con_opt}, we normalize $\boldsymbol{\lambda}$ to the unit ball and thus the second assumption is satisfied.

\section{Experiments}

\begin{table*}[t]
\caption{
\small
Performance of local masks.
The best methods except Pick-all on each dataset is boldfaced,
and the runner-up is highlighted by $\ast$.
$\circ$ indicates significantly better performance according to $t$-tests.
Column Pick-all provides lower (underlined) and upper bounds (overlined) of faithfulness
(the counterfactual (CF) of selecting all features is equivalent to the faithfulness (FA) of selecting no feature)).
%
%
}
\label{tab:overall}
\centering
\scriptsize

\subcaption*{\small
Performance (mean with std. in parenthesis) of each algorithm in terms of
\textit{faithfulness}. 
Lower is better ($\downarrow$).}
\begin{tabular}{ c | l l l | l l | l l l l l}
\toprule
\textbf{Method} &  Pick-all 
& DES & SNX-global & SM & SNX-UC & SNX-KL & SNX-DES & SNX-inter & SNX-union & SNX \\
\midrule
Adult & \underline{0.65} (0.14)
& 1.67 (2.46) & 0.71 (0.18) & 0.81 (0.72) & 0.68 (0.14) $\ast$  & 0.69  (0.20) & \textbf{0.66} (0.14) & \textbf{0.66} (0.14) & \textbf{0.66}  (0.14) & \textbf{0.66} (0.14) $\circ$ \\
Bank & \underline{0.62}  (0.16)
& 0.82  (0.85) & 0.72  (0.22) & 0.71  (0.24) & 0.65 (0.16) $\ast$ & 0.66  (0.16) & \textbf{0.63}  (0.16) & \textbf{0.63}  (0.16) & \textbf{0.63}  (0.16) & \textbf{0.63}  (0.16) $\circ$ \\
Credit & \underline{0.64}  (0.15)
& 1.15 (1.73) & 0.70  (0.19) & 1.13  (1.35) & 0.72  (0.19) & 0.74 (0.21) & 0.69 (0.18) $\ast$ & \textbf{0.68}  (0.16) & \textbf{0.68}  (0.16) & \textbf{0.68}  (0.16) $\circ$ \\
COMPAS & \underline{0.62}  (0.19)
& 0.76  (0.57) & 0.71 (0.65) & 0.79  (0.73) & 0.64 (0.17) & 0.63 (0.19) $\ast$ & \textbf{0.62}  (0.19) & \textbf{0.62}  (0.19) & \textbf{0.62}  (0.19) &
\textbf{0.62}  (0.19) $\circ$ \\
\bottomrule
\end{tabular}

\subcaption*{\small
Performance (mean with std. in parenthesis) of each algorithm in terms of
\textit{counterfactual}. 
Higher is better. ($\uparrow$)}
\begin{tabular}{ c | l l l | l l | l l l l l}
\toprule
\textbf{Method} &  Pick-all 
& DES & SNX-global & SM & SNX-UC & SNX-KL & SNX-DES & SNX-inter & SNX-union & SNX \\
\midrule
Adult & $\overline{7.50}$ (2.05)
& 0.85  (0.49) & 0.78 (0.14) & 1.02 (1.09) & 1.28 (1.57) $\ast$ & 0.99  (0.92) &\textbf{1.31} (1.62) $\circ$ & 1.25 (1.53) & 1.26 (1.54) & 1.26 (1.54) \\ 
Bank & $\overline{6.86}$ (2.49)
& 1.26 (1.58) & 0.71 (0.11) & 1.67 (2.09) & 1.80 (2.25) & 1.25 (1.47) & 2.54 (2.83) & 2.78 (2.97) $\ast$ & 2.78 (2.97) $\ast$ & \textbf{2.79}  (2.97) $\circ$ \\
Credit & $\overline{7.54}$ (2.26) 
& 0.74  (0.23) & 0.83 (0.19) & 1.23 (0.82) $\ast$ & \textbf{1.24} (1.17) $\circ$ & 0.89 (0.57) & 1.16 (1.10) & 0.97 (0.90) & 0.97 (0.90) & 0.97 (0.88) \\
COMPAS & $\overline{7.02}$ (2.52)
& 1.86 (2.53) & 0.72 (0.09) & 1.76 (2.07) & 3.65 (3.22) & 2.65 (2.74) & 4.56 (3.48) & 4.93 (3.43) & \textbf{5.00} (3.43) $\circ$ &
4.98 (3.43) $\ast$ \\
\bottomrule
\end{tabular}



\end{table*}

\begin{table}
\footnotesize
\caption{
\small
Performance of global invariant masks in \textit{faithfulness} (FA) and \textit{counterfactual} loss (CF), reported in 
mean with std. in parenthesis.
The best methods except \textit{Pick-all} are boldfaced.
The underlines for FA and overlines for CF in the column ``Pick-all''
indicate the lower and upper bounds of FA (the CF of selecting all features is equivalent to the FA of selecting no feature). 
Sensitivity study w.r.t. different numbers of selected features is given in Fig. \ref{fig:sensitivity}.
%
}
\label{tab:overall-global}
\centering
\begin{tabular}{ 
c || c |
c c c 
}
\toprule
    \textbf{Metrics} &
\textbf{Datasets} & Pick-all 
& DES
& SNX-global \\
\midrule
\multirow{6}{*}{\textbf{FA} ($\downarrow$)} 
& Adult & \underline{0.00} (0.00)
& 0.28  (0.28)
& \textbf{0.00}  (0.00) \\ 
& Bank & \underline{0.00}  (0.00)
& 0.21  (0.22) 
& \textbf{0.00}  (0.00) \\
& Credit & \underline{0.00}  (0.00) 
&  0.31  (0.36) 
& \textbf{0.00}  (0.00) \\
& COMPAS & \underline{0.00}  (0.00) 
& 0.11  (0.19) 
& \textbf{0.00}  (0.00) \\ 
\cmidrule{2-5}
& Molecule & \underline{0.08} (0.00) 
& - & \textbf{0.91} (1.01) \\
& BP & \underline{0.01} (0.00) 
& - & \textbf{0.00} (0.00) \\
\midrule
\multirow{6}{*}{\textbf{CF} ($\uparrow$)} 
& Adult & $\overline{0.69}$ (0.21) 
& 0.33 (0.21) 
& \textbf{0.69}  (0.21) \\
& Bank & $\overline{0.62}$ (0.26) 
& 0.34 (0.19) 
& \textbf{0.62}  (0.26) \\
& Credit & $\overline{0.62}$ (0.21) 
&  0.31 (0.26) 
&  \textbf{0.63} (0.21) \\
& COMPAS & $\overline{0.56}$ (0.18) 
& \textbf{0.56} (0.21)
& \textbf{0.56}  (0.18) \\
\cmidrule{2-5}
& Molecule & $\overline{6.80}$ (0.01)
& - & \textbf{6.61} (0.49) \\
& BP & $\overline{1.11}$ (1.39) 
& - & \textbf{1.22} (1.54) \\
\bottomrule

\end{tabular}
\end{table}

\subsection{Experimental settings}
\noindent\textbf{Datasets.}
We conduct experiments on four tabular datasets~\cite{moro2014data,mothilal2020explaining} and two graph datasets (molecules~\cite{paragian2020computational} and brain networks~\cite{ma2019deep}).
We split the instances into training ($70\%$) and test ($30\%$) sets, 
and Section \ref{sec:repro-list} provides details about pairs generation.
The target SN is trained on the training portion and then held fixed during explanation generation on the test portion.


\noindent\textbf{Metrics.}
We evaluate the faithfulness (FA) by Eq. (\ref{eq:faith}), counterfactual loss (CF) by Eq. (\ref{eq:cf}), and conformity of the explanations with the same percentage of selected salient features or edges.
We also study the sensitivity of these metrics w.r.t. a different number of selected features/edges (simplicity).
The conformity is evaluated by averaging the Jaccard similarity between global and local masks,
$J = \frac{ | a(\textbf{M}) \cap a(\textbf{m}) | }{| a(\textbf{M}) \cup a(\textbf{m}) | }$,
between $\mathbf{M}^s$ and $\mathbf{m}^s$, and between $\mathbf{M}^t$ and $\mathbf{m}^t$, respectively.


\noindent\textbf{Baselines and variants.}
There are several options to obtain local masks $\mathbf{m}^s$ for the queries.
The following three methods use global 
masks as local masks and are agnostic to references.
\begin{itemize}[leftmargin=*]
    \item Pick-all: set all elements in the global masks to one.
    \item DES (tabular data only): an unsupervised feature selection~\cite{wei2017rethinking} that generates
    pseudo labels for all pairs of instances using kNN to supervise the learning of a feature selector.
    The selected features of the query are used when comparing with different references.
    \item SNX-global: solve Eq. (\ref{eq:obj_gm}) for $\mathbf{M}^s$ which is treated as a local mask without stage 2 local mask optimization.
\end{itemize}

\noindent The following baselines disregard constraints by global masks.
\begin{itemize}[leftmargin=*]
    \item 
    Saliency maps (SM)~\cite{baldassarre2019explainability}: take the gradient of $\ell^{\textnormal{SN}}$ with respect to the input $\mathbf{x}^s$ and retain the features with the largest gradient magnitudes.
    \item 
    SNX-unconstrained (SNX-UC): 
    minimize the objective Eq. (\ref{eq:con_lm_obj})
    without the constraints Eqs. (\ref{eq:con_lm_con_1})-(\ref{eq:con_lm_con_2}).
    \item 
    GNNExplainer (GNNExp)~\cite{ying2019gnnexplainer}: 
    learn soft masks for edges by maximizing the mutual information (MI) between the model's predictions w.r.t. original and masked graphs,
    and use the entropy as part of regularization to encourage the sparsity and simplicity of masks.
    Applying GNNExp~\cite{ying2019gnnexplainer} to SN is similar to SNX-UC,
    but with different objective functions and regularization terms in Eq. (\ref{eq:con_lm_obj}).
    \item
    PGExplainer (PGExp)~\cite{luo2020parameterized}: 
    train a shareable generator for all graphs by maximizing the MI between the SN predictions on the original and masked graphs.
    The generator generates edge masks for each graph to be explained.
\end{itemize}

\noindent
The following three baselines extract local masks with variants of global masks.
They apply to tabular data only the features are aligned in the query and the reference.
\begin{itemize}[leftmargin=*]
    
    \item
    SNX-DES: optimize $\mathbf{m}$ by 
    Eqs. (\ref{eq:con_lm_obj})-(\ref{eq:con_lm_con_2}),
    with global 
    mask $\mathbf{M}^s$ found by DES used in the constraints Eqs. 
    (\ref{eq:con_lm_con_1})-(\ref{eq:con_lm_con_2}).
     \item SNX-inter:
    a variant of SNX. Instead of using $\mathbf{M}^s$ to constrain $\mathbf{m}^s$, this baseline uses the element-wise minimum
    $\mathbf{M} = \min (\mathbf{M}^s, \mathbf{M}^t)$, analogous to intersecting the two global masks over the query and the reference, leading to a more rigid global mask constraint.
    \item SNX-union: similar to SNX-inter but use  the element-wise maximum
    $\mathbf{M}=\max (\mathbf{M}^s,  \mathbf{M}^t)$ to simulate the union of $\mathbf{M}^s$ and $\mathbf{M}^t$,
    leading to a less rigid constraint.
\end{itemize}

\subsection{Quantitative evaluation on tabular datasets}
\label{sec:exp_global_mask}

\noindent\textbf{Evaluations of global masks.}
We use $\ell(f(\mathbf{x}, \mathbf{x}), f(\mathbf{x}, \mathbf{M} \otimes \mathbf{x}))$ as faithfulness and $\ell(f(\mathbf{x}, \mathbf{x}), f(\mathbf{x}, (1-\mathbf{M}) \otimes \mathbf{x}))$ as CF loss to evaluate global masks. 
By default, we select the top 10 most important minor features into the global masks.
In Table \ref{tab:overall-global},
with all features selected,
the baseline Pick-all
does not generate meaningful explanations
but provides the best performance in FA and CF, respectively (indicated by the under- and over-lines).
SNX-global outperforms DES and achieves the same performance as Pick-all.
Since all samples in the four tabular datasets have less than 10 major features,
when SNX-global can pick 10 minor features,
self-supervised learning (Eq. (\ref{eq:obj_gm})) can consider the target SN and select the active locations in the one-hot vectors of \textit{each} instance.

\vspace{.05in}
\noindent\textbf{Evaluations of local masks.}
In Table~\ref{tab:overall},
we compare the faithfulness 
and counterfactual losses
of local masks found by various methods.
We answer the following questions.

\vspace{.05in}
\noindent\textit{Does the two-stage optimization find better local masks?}
Overall, the best-performing methods in local mask faithfulness are in the last four columns representing variants of the SNX using GDA, with different global masks as constraints.
In terms of counterfactual loss, 
the optimal local masks outperform the remaining methods,
except on the Credit dataset (SNX-UC has no constraint and can include more salient features).

\vspace{.05in}
\noindent\textit{Which variants of global masks help} SNX?
We further compare the last four columns of Table~\ref{tab:overall} and find similar faithfulness losses.
However, there is no clear winner in terms of counterfactual loss.
Since we aim to interpret a local reference-specific explanation along with the associated global invariant mask as a context,
SNX and its variants are preferred over DES,
based on Table~\ref{tab:overall-global}.

\begin{figure}[t]
    \centering
    \includegraphics[width=0.42\textwidth]{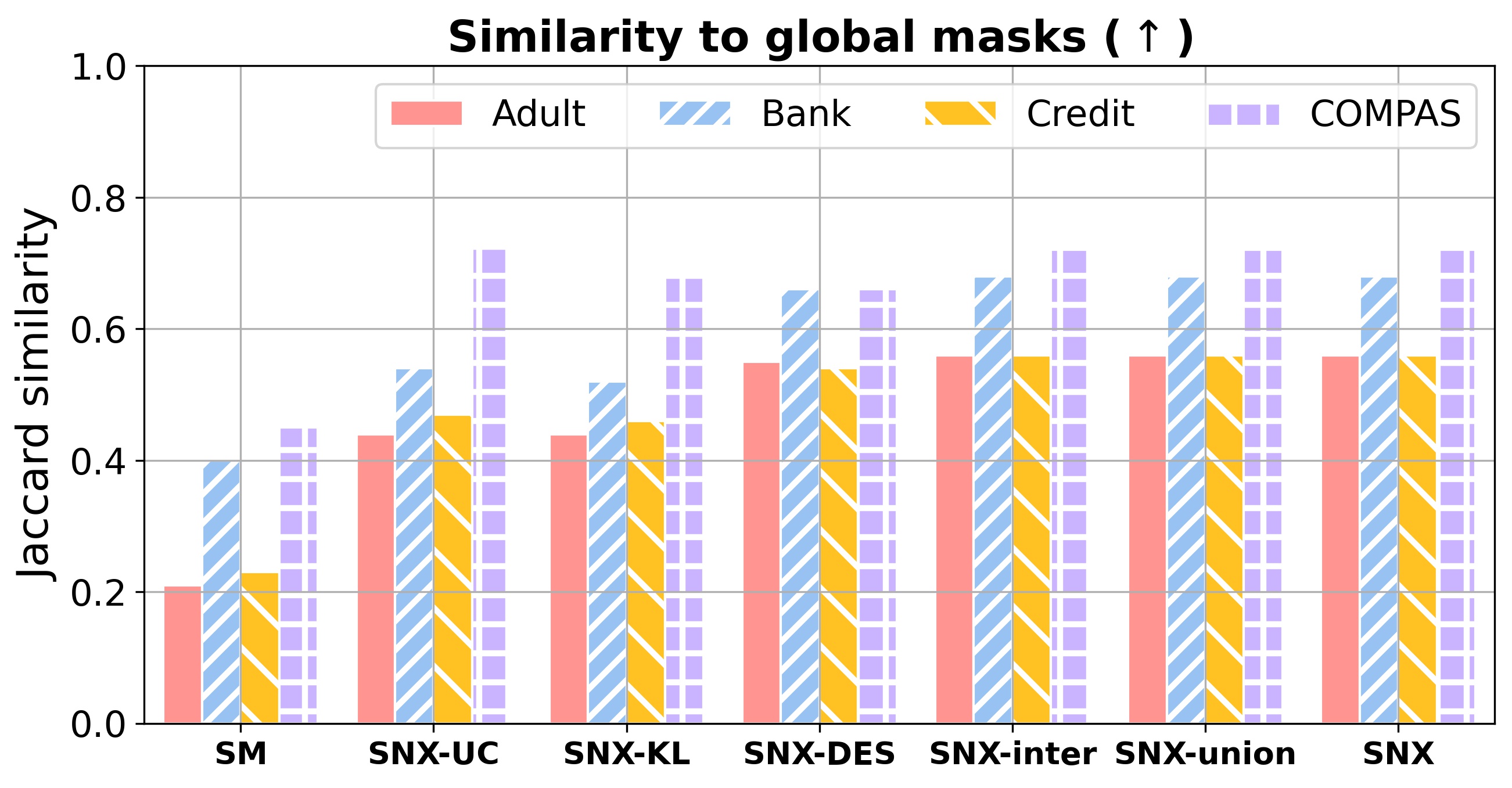}
    \caption{
    \small Jaccard similarity measuring conformity of multiple local masks to a global invariant mask. 
    %
    %
    }
    \label{fig:tabular_distance_to_global}
\end{figure}

\vspace{.05in}
\noindent\textit{Why do the other methods underperform SNX and its variants?}
\begin{itemize}[leftmargin=*,noitemsep,topsep=0pt]
    \item 
DES does not perform well 
and has the worst faithfulness except on COMPAS, where SM is the worst.
That is because DES is agnostic about the SN architecture and
does not consider the reference instance when finding $\mathbf{M}^s$ to mask both $\mathbf{x}^s$ and $\mathbf{x}^t$.
DES is also not performing well in the counterfactual loss,
indicating that it fails to include the most salient features.
These drawbacks can be addressed by the stage 2 optimization where the SN model is put back (SNX-DES is among of the best performers, noting that the global masks found by DES are only soft constraints).
\item
SNX-global
uses $\mathbf{M}^s$ found by the self-supervised mask learning and takes the target SN into account.
As a result,
it has good performance in faithfulness.
However, it is worse than DES in the counterfactual loss on 3 out of 4 datasets, meaning that it can miss even more salient features.
\item
SM takes the reference into account and
outperforms DES in both metrics except on COMPAS.
Interestingly, on the Credit dataset,
this method outperforms those methods in the last four columns in the counterfactual loss,
indicating that gradients can locate the most salient features.
\item
SNX-UC: without constraints from the global masks,
it performs stage 2 optimization to find local masks.
Therefore, it is a strong baseline:
it is the runner-up in faithfulness on Adult and Bank and is the best in counterfactual loss on Credit.
However, no constraint leads to less conformity to global invariant masks (Fig.~\ref{fig:tabular_distance_to_global} and Table~\ref{tab:simplified_case_study}).
\item
SNX-KL is very similar to SNX-UC, except that the soft constraints are implemented as KL-divergence penalty terms.
They have similar faithfulness performance, but SNX-KL significantly underperforms SNX-UC in counterfactual loss, indicating that the constraints are actually working by selecting just enough salient features but possibly excluding redundant salient ones.
\end{itemize}

\vspace{.05in}
\noindent\textit{How the local masks conform to the global invariant masks?}
In Fig.~\ref{fig:tabular_distance_to_global},
we report the average of Jaccard similarities between the important major features selected by global masks $\textbf{N}^s$ and local masks $\textbf{n}^s$, 
when comparing $\mathbf{x}^s$ to multiple $\mathbf{x}^t$.
The higher similarity indicates the more conformity of local masks to global masks.
Moreover,
high conformity indicates that local masks of a query w.r.t. different references are regulated by global invariant masks, and
these local masks are more robust, 
since they highlight different features adaptively within the selected global invariant features, when comparing with various references.
As shown in Fig.~\ref{fig:tabular_distance_to_global},
except for the COMPAS dataset, SNX with GDA results in the best conformity (highest similarity), 
regardless of the global invariant masks used.
SNX is the runner-up on COMPAS next to SNX-UC.
The low CF of SNX-global in Table \ref{tab:overall} can be blamed.
SNX-global can find some important features but miss some as well.
SNX bypass global constraints and find other important features missed by SNX-global for better FA.
In Section \ref{sec:conformity}, we further study the conformity.

\subsection{Quantitative evaluation on graph datasets}
\begin{figure}[t]
    \centering
\begin{minipage}{.42\textwidth}
    \includegraphics[width=\textwidth]{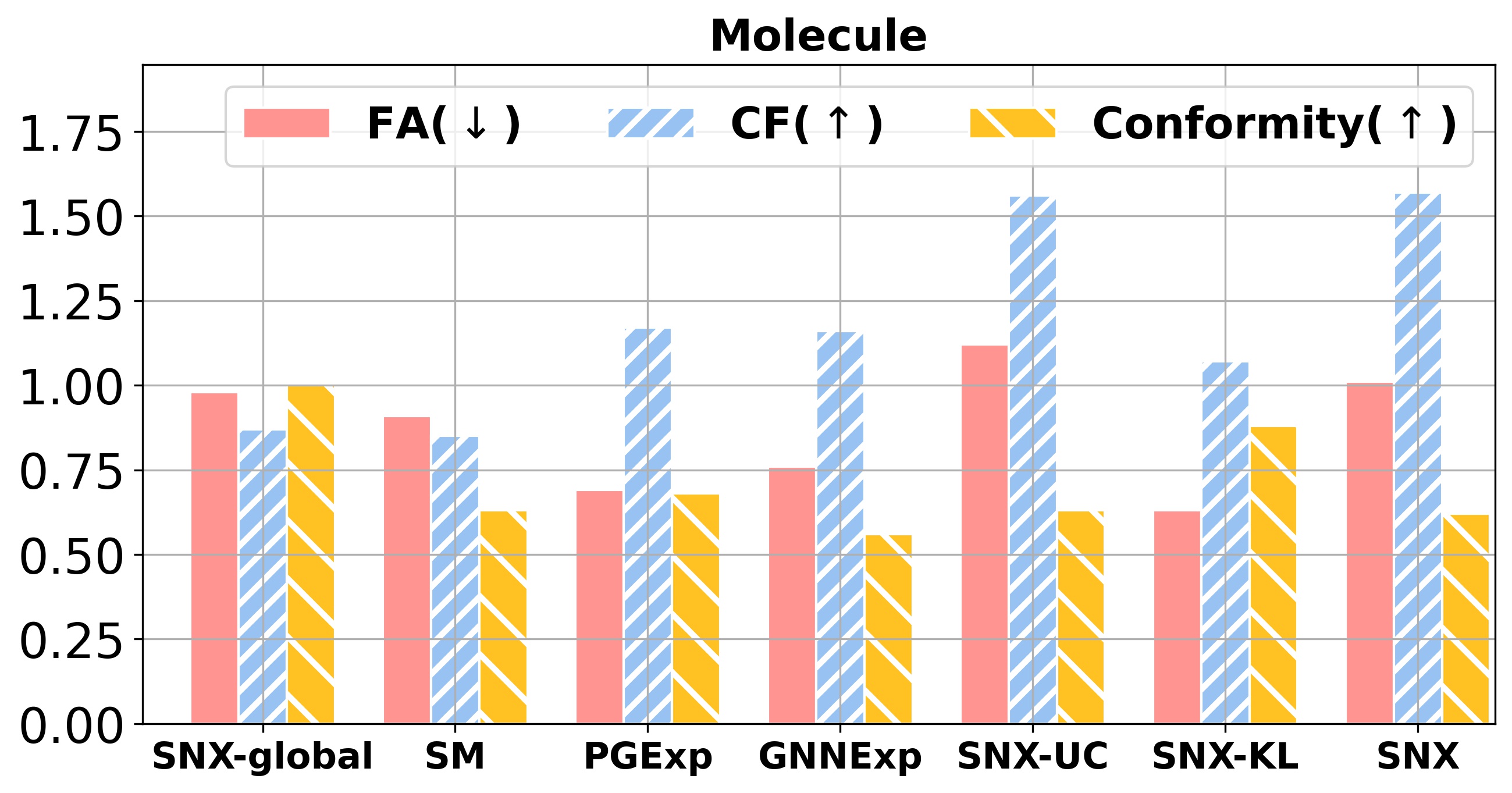}
\end{minipage}
\\
\begin{minipage}{.42\textwidth}
    \includegraphics[width=\textwidth]{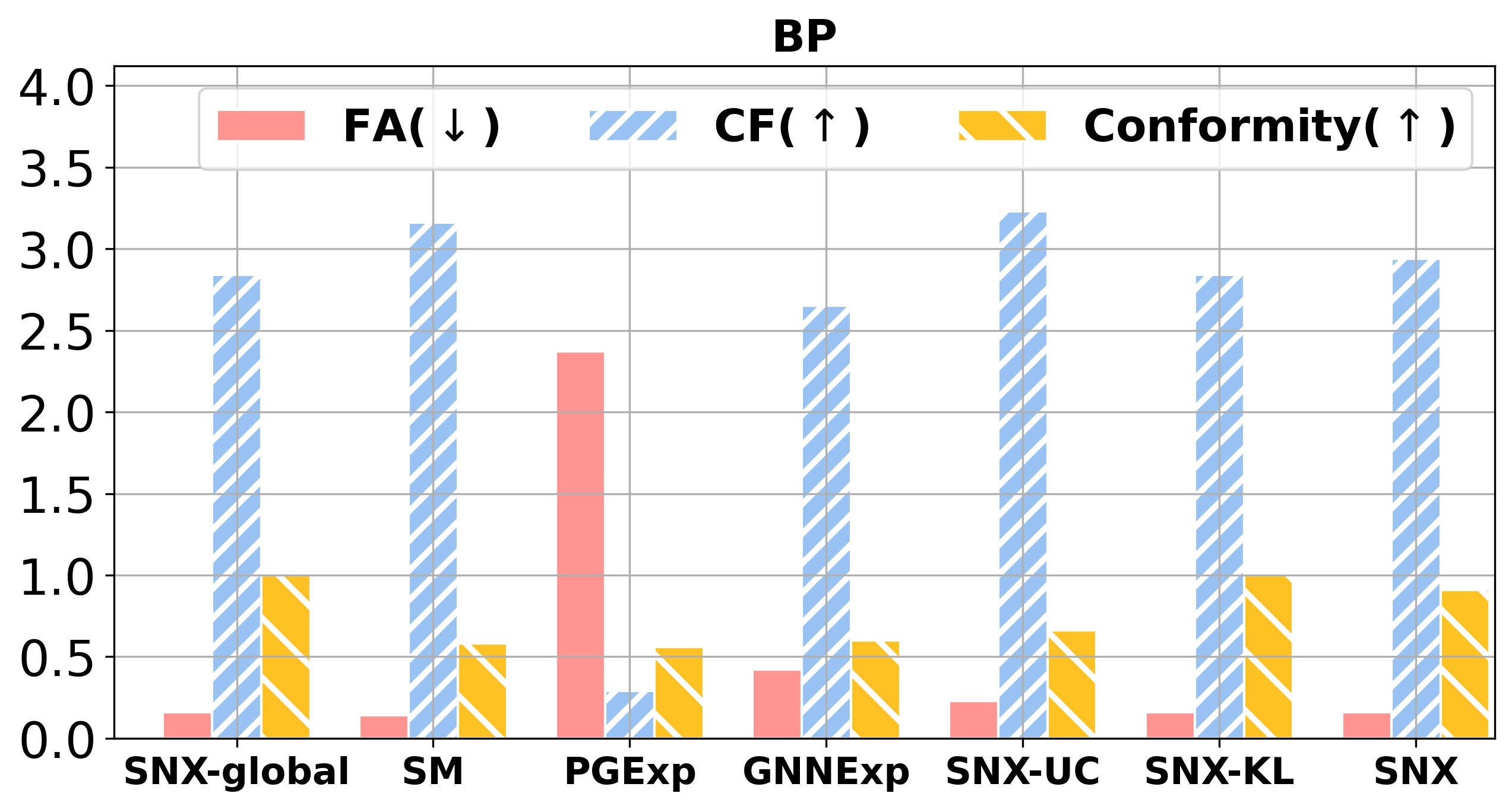}
\end{minipage}%
    \caption{\small Faithfulness (FA), counterfactual (CF), and conformity (in Jaccard similarity) of local masks on graph datasets.
    %
    %
    }
    \label{fig:graphs_local_overall}
\end{figure}


For graph data, we select the top 75\% important edges by default.
In Table~\ref{tab:overall-global},
we evaluate the average faithfulness and counterfactual loss of the global masks over all test graphs.
SNX-global works well on graph data, with 
mean faithfulness much closer to the lower bound than the upper bound (represented by FA and CF in the Pick-all column).

In Fig.~\ref{fig:graphs_local_overall}, 
we compare local graph masks with respect to faithfulness, counterfactual loss, and conformity.
On the molecule dataset,
SNX-KL can achieve the best (the lowest) faithfulness and highest conformity ($\sim$90\%) to global masks found by SNX-global (which always has conformity 1).
SNX with GDA  and SNX-UC have the best CF loss,
while SNX achieves better (lower) FA than SNX-UC.
Both GNNExp and PGExp are competitive baselines, 
whose counterfactual is similar to SNX-KL's but are significantly worse in the other two metrics.
Four methods, SM, GNNExp, PGExp, and SNX-UC, disregard any constraints and have worse conformity
(overlapping only about 60\% of global masks' edges)
and worse faithfulness than SNX-KL.

On the BP dataset where graphs are much larger,
both SNX-KL and SNX achieve better FA and conformity, and thus more robust than other methods.
Since SNX-global has already provided very competitive global masks w.r.t. FA and CF, 
SNX-KL can result in a similarly good performance when following SNX-global a lot,
while SNX sacrifices a bit of conformity to bypass some global constraints for better masks w.r.t. FA and CF.
On the contrary,
SM has the worst conformity (<50\%) with only a slightly better faithfulness than SNX-KL,
and it validates the analysis in Section \ref{sec:xml} that SM has worse robustness 
as it is too sensitive to different references given a query.
PGExp performs poorly on all three metrics,
while GNNExp is much better (but still worse than SNX-KL and SNX in all three metrics).
The reason can be that all pairs of graphs share the same generator in PGExp while SNX-based methods and GNNExp optimize masks for each specific pair of graphs.
Compared with molecule dataset, 
graphs in BP are much larger and it is harder to find shared parameters that work for all pairs of graphs.

\subsection{Convergence of GDA}
\label{sec:convergence}
\begin{figure}[t]
    \centering
\begin{minipage}{.24\textwidth}
    \includegraphics[width=\textwidth]{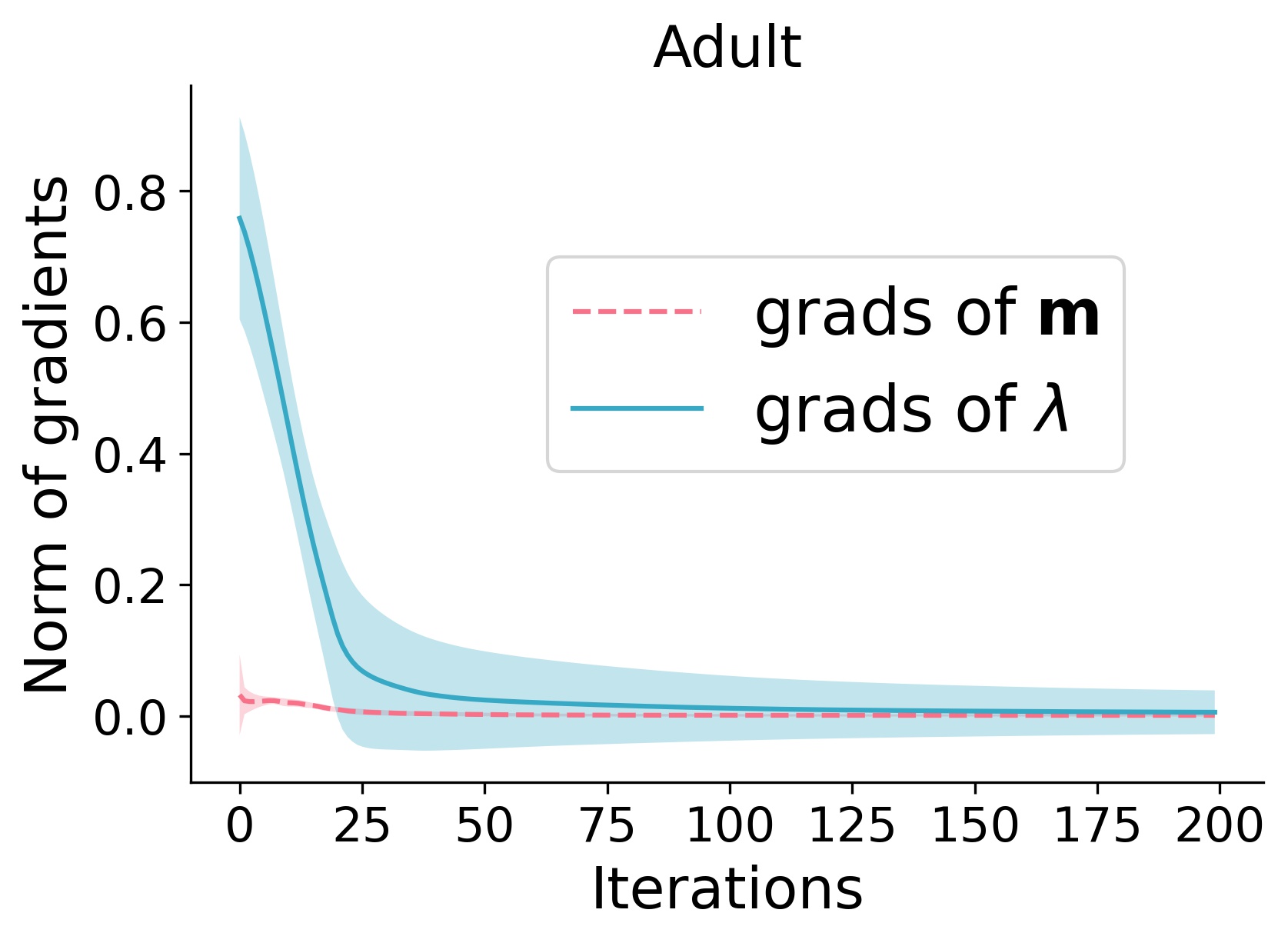}
\end{minipage}%
\begin{minipage}{.24\textwidth}
    \includegraphics[width=\textwidth]{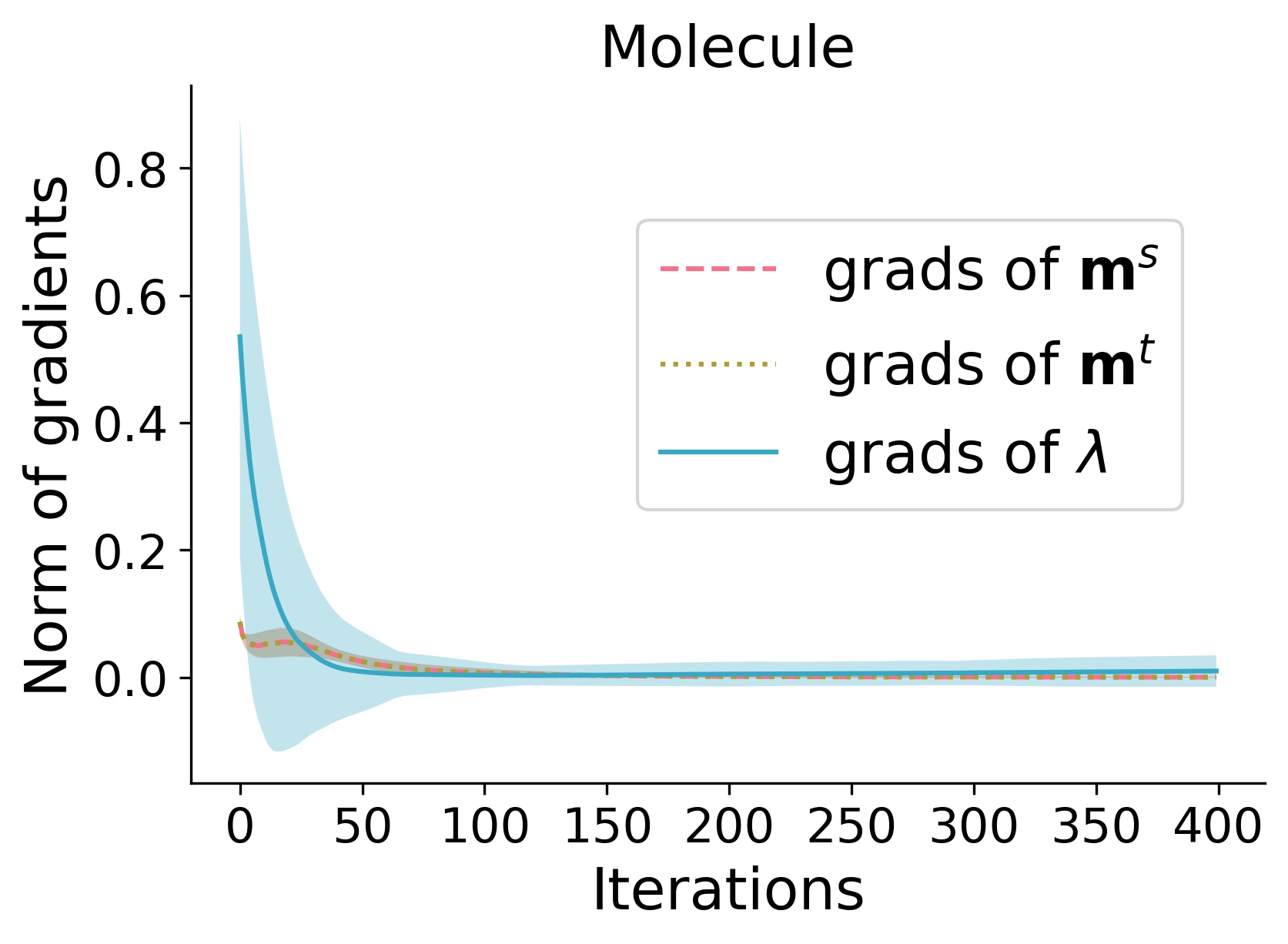}
\end{minipage}%
    \caption{\small $\ell2$ norms of the gradients of Lagrangian w.r.t. $\mathbf{m}$'s and $\boldsymbol{\lambda}$'s when running GDA on the Adult and Molecule datasets. 
    Line indicates the mean and the colored region for variance.
    }
    \label{fig:grads}
\end{figure}


To track the convergences of GDA,
Fig. \ref{fig:grads} records the norm of the gradients of the Lagrangian w.r.t. the primal and dual variables.
Both norms go close to 0 in 100 iterations,
confirming the convergence of GDA.
Similar trends are found in other datasets (not shown due to space limit).

\subsection{More on conformity}
\label{sec:conformity}

\begin{figure}[t]
    \centering
\begin{minipage}{.16\textwidth}
    \includegraphics[width=\textwidth]{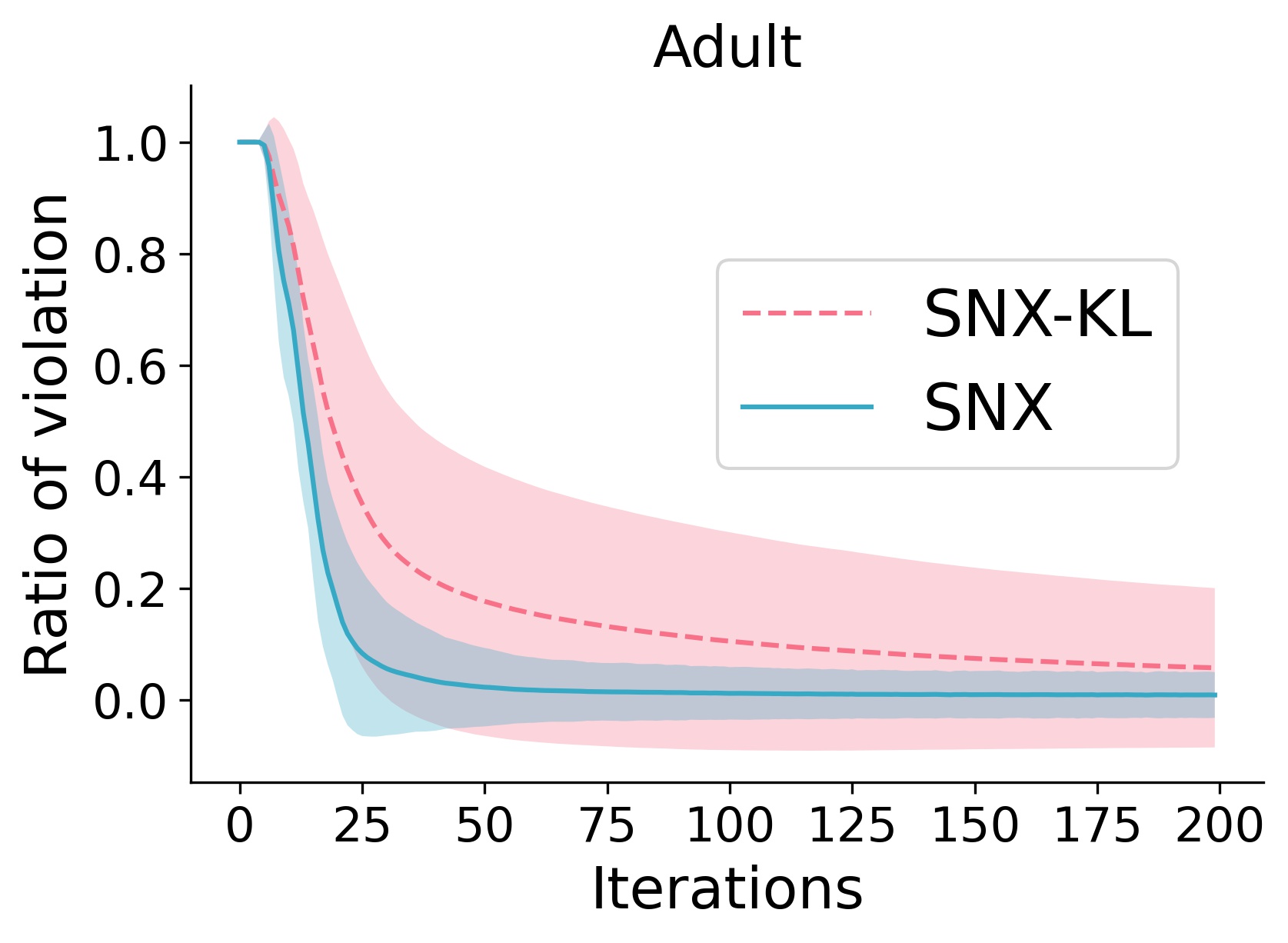}
\end{minipage}%
\begin{minipage}{.16\textwidth}
    \includegraphics[width=\textwidth]{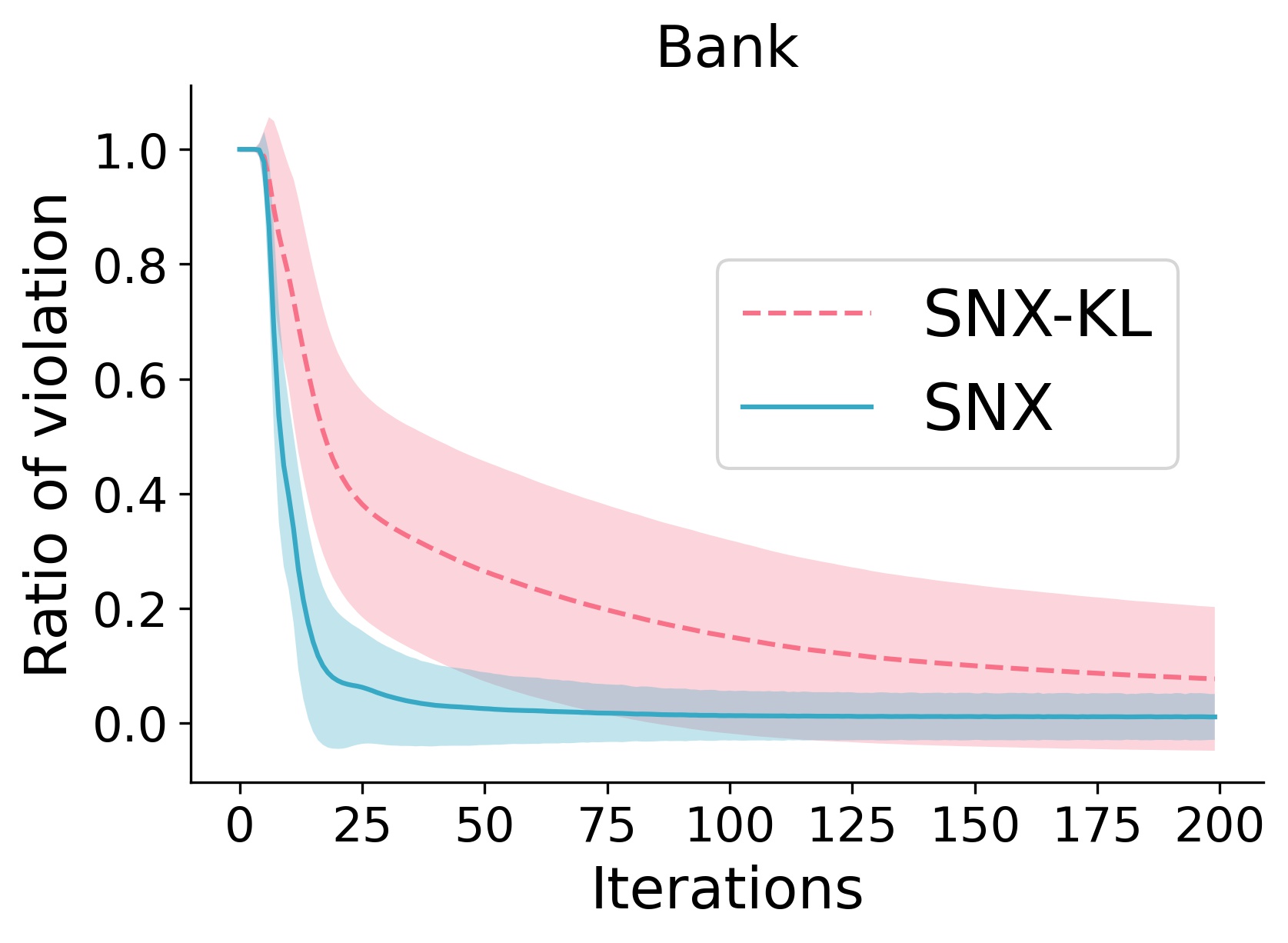}
\end{minipage}%
\begin{minipage}{.16\textwidth}
    \includegraphics[width=\textwidth]{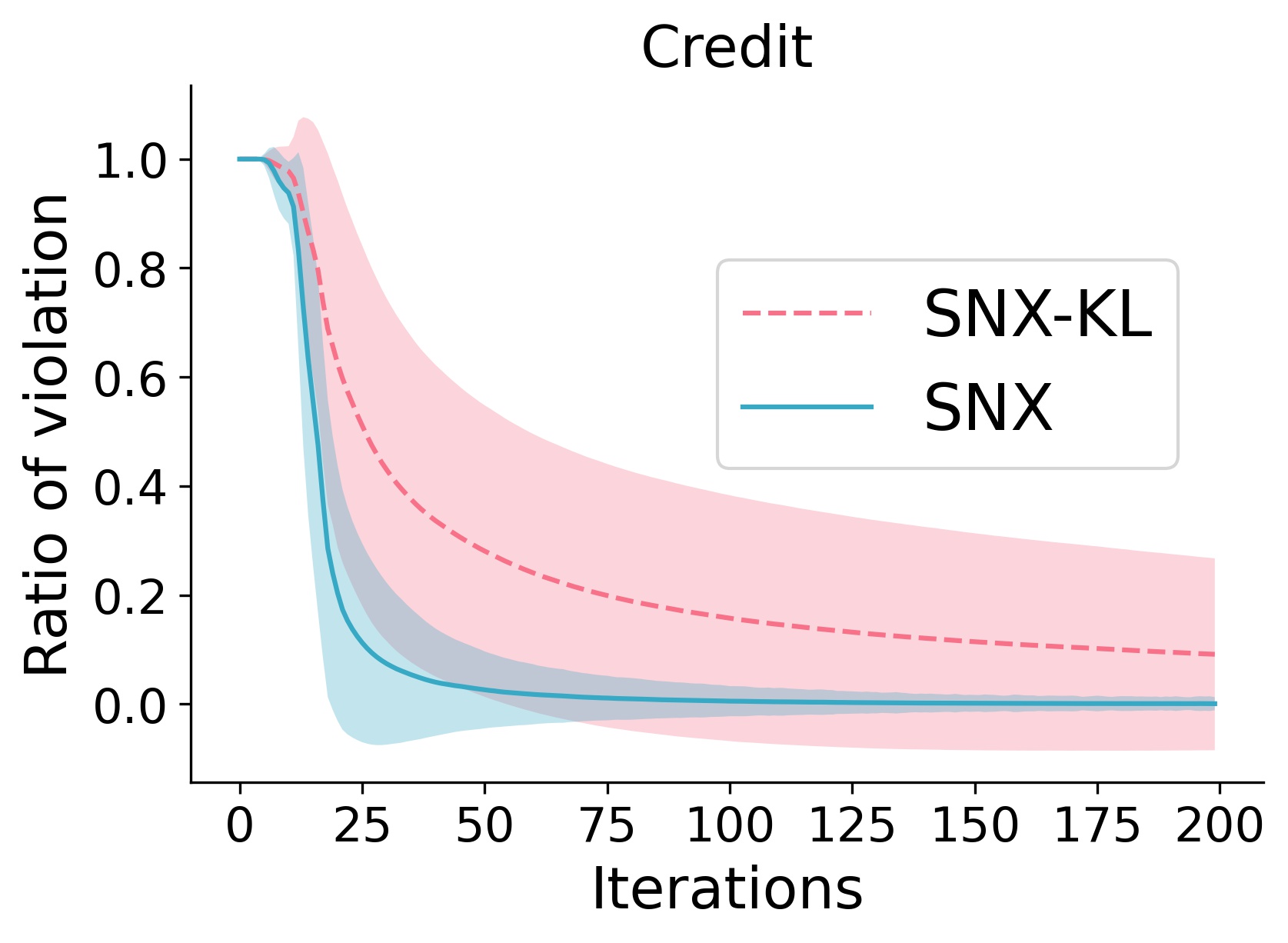}
\end{minipage}\\
\begin{minipage}{.16\textwidth}
    \includegraphics[width=\textwidth]{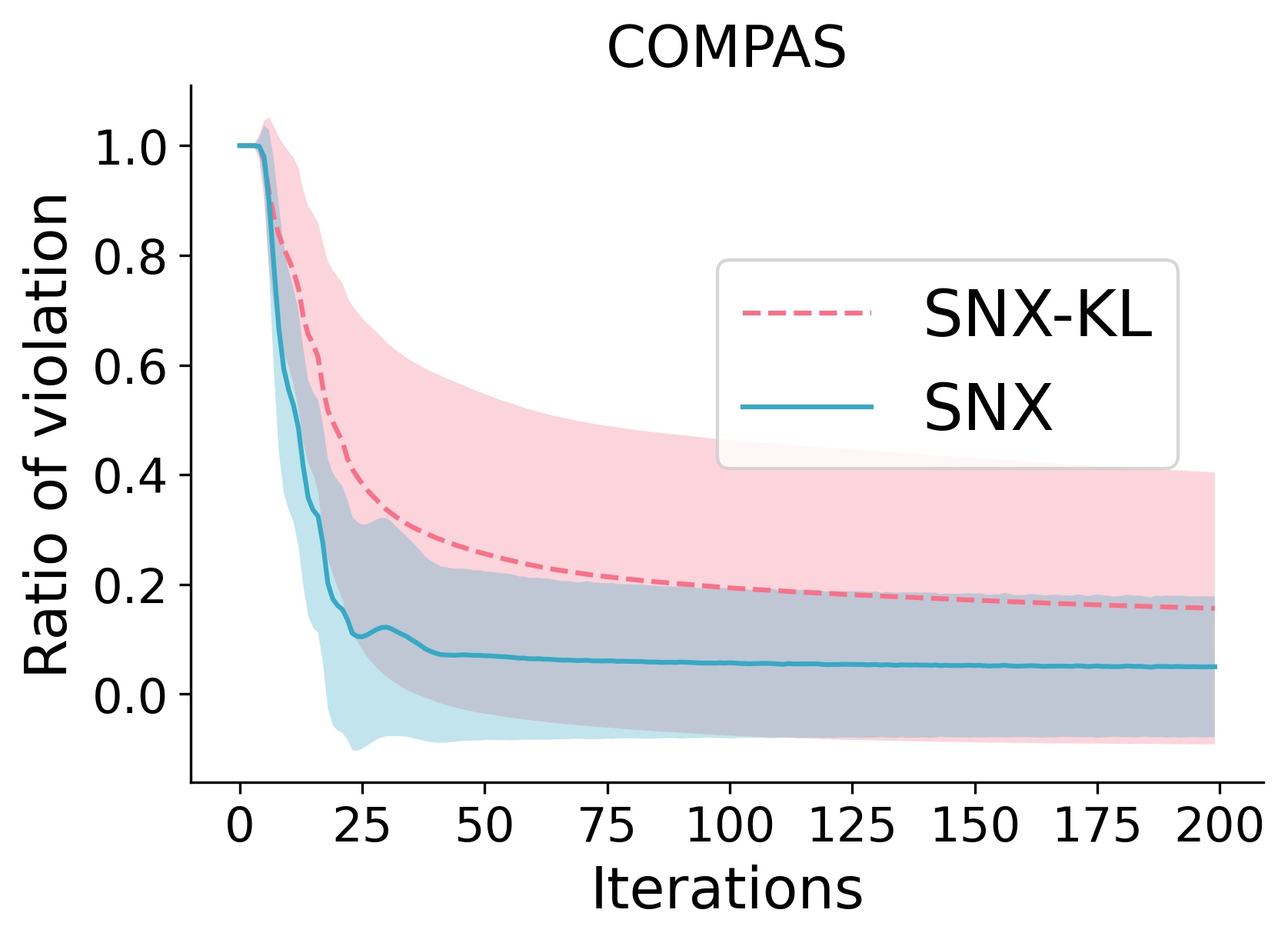}
\end{minipage}%
\begin{minipage}{.16\textwidth}
    \includegraphics[width=\textwidth]{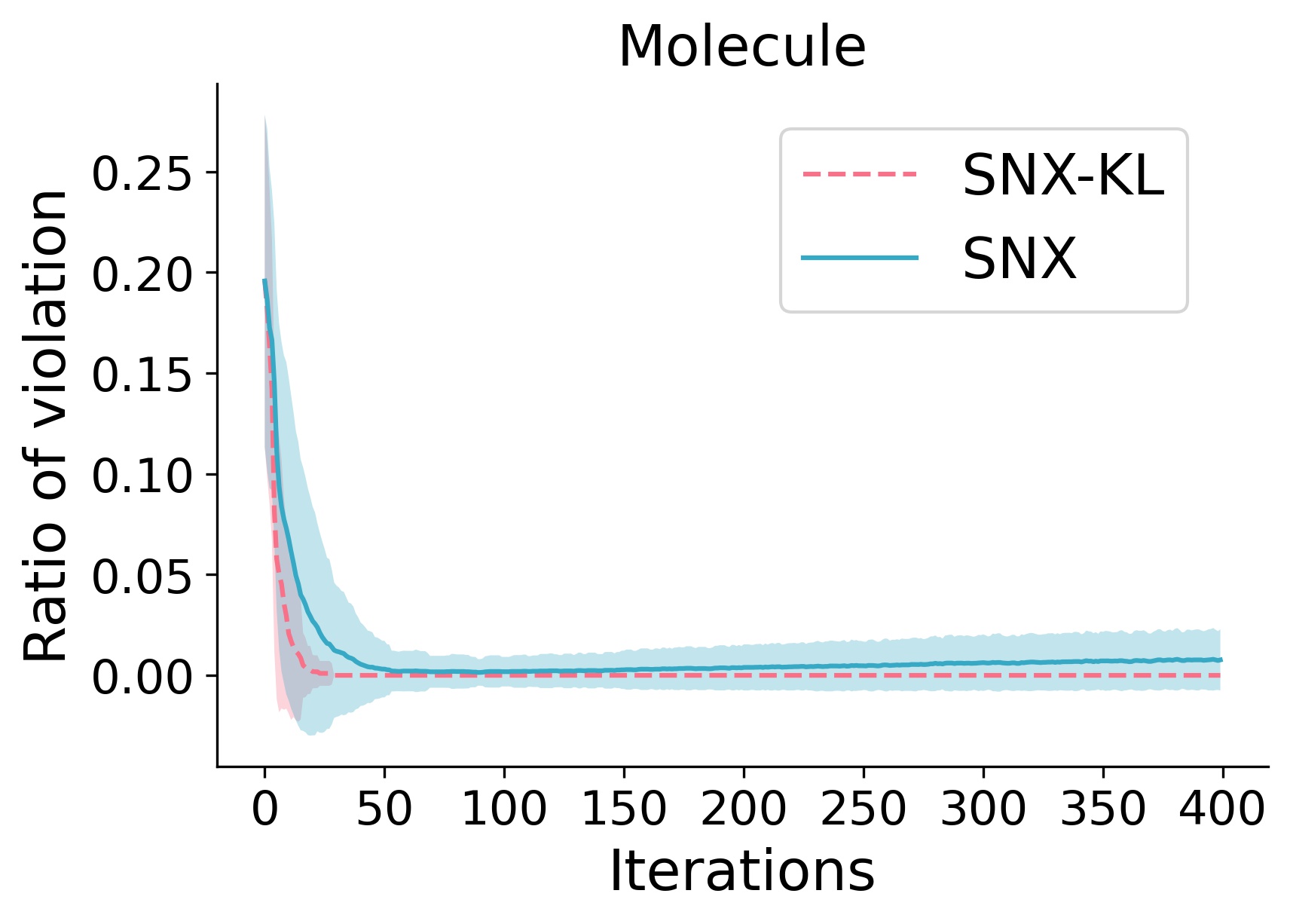}
\end{minipage}%
\begin{minipage}{.16\textwidth}
    \includegraphics[width=\textwidth]{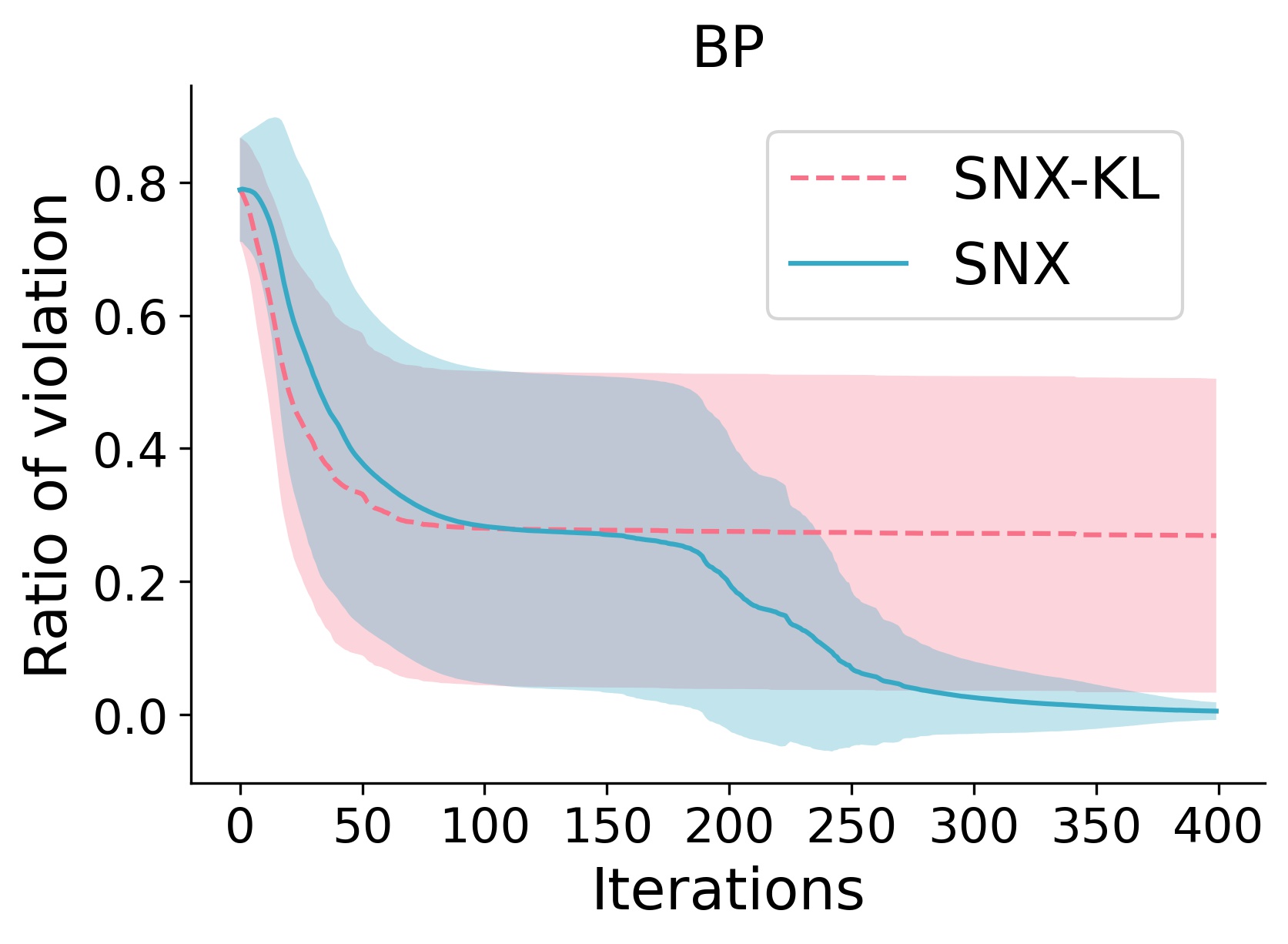}
\end{minipage}
    \caption{\small The ratio of instances violates constraints in the first 200 (tabular data) and 400 iterations (graph data) by SNX-KL and SNX.
    }
    \label{fig:kl-lag}
\end{figure}

Fig. \ref{fig:kl-lag} 
compares the violation of constraints by SNX and SNX-KL during optimization for all six datasets.
We evaluate Eqs. (\ref{eq:con_lm_con_1})-(\ref{eq:con_lm_con_2}) using the masks without selecting top edges.
To make both methods comparable,
we set the same hyper-parameters as illustrated in Section \ref{sec:repro-list}
using no pre-train to ensure that SNX-KL and SNX start with the same initialization.
SNX breaks fewer constraints in four tabular datasets and BP.
Notice that SNX-KL keeps violating about 30\% of constraints after about 70 iterations.
We conjecture that SNX-KL converges to a ``minimizer'' balancing both the objective functions and soft constraints.
SNX-KL has higher violations in Fig. \ref{fig:kl-lag} but better conformity in Fig. \ref{fig:graphs_local_overall}.
since the top 75\% edges are selected from the masks to generate Fig. \ref{fig:graphs_local_overall}, 
and masks are evaluated directly in Fig. \ref{fig:kl-lag}.
The decoding may select one edge but drop the other even if their importance are the same. 

\subsection{Sensitivity analysis}
\label{sec:sensitivity_analysis}
\begin{figure}[t] 
    \centering
\begin{minipage}{.2\textwidth}
    \includegraphics[width=\textwidth]{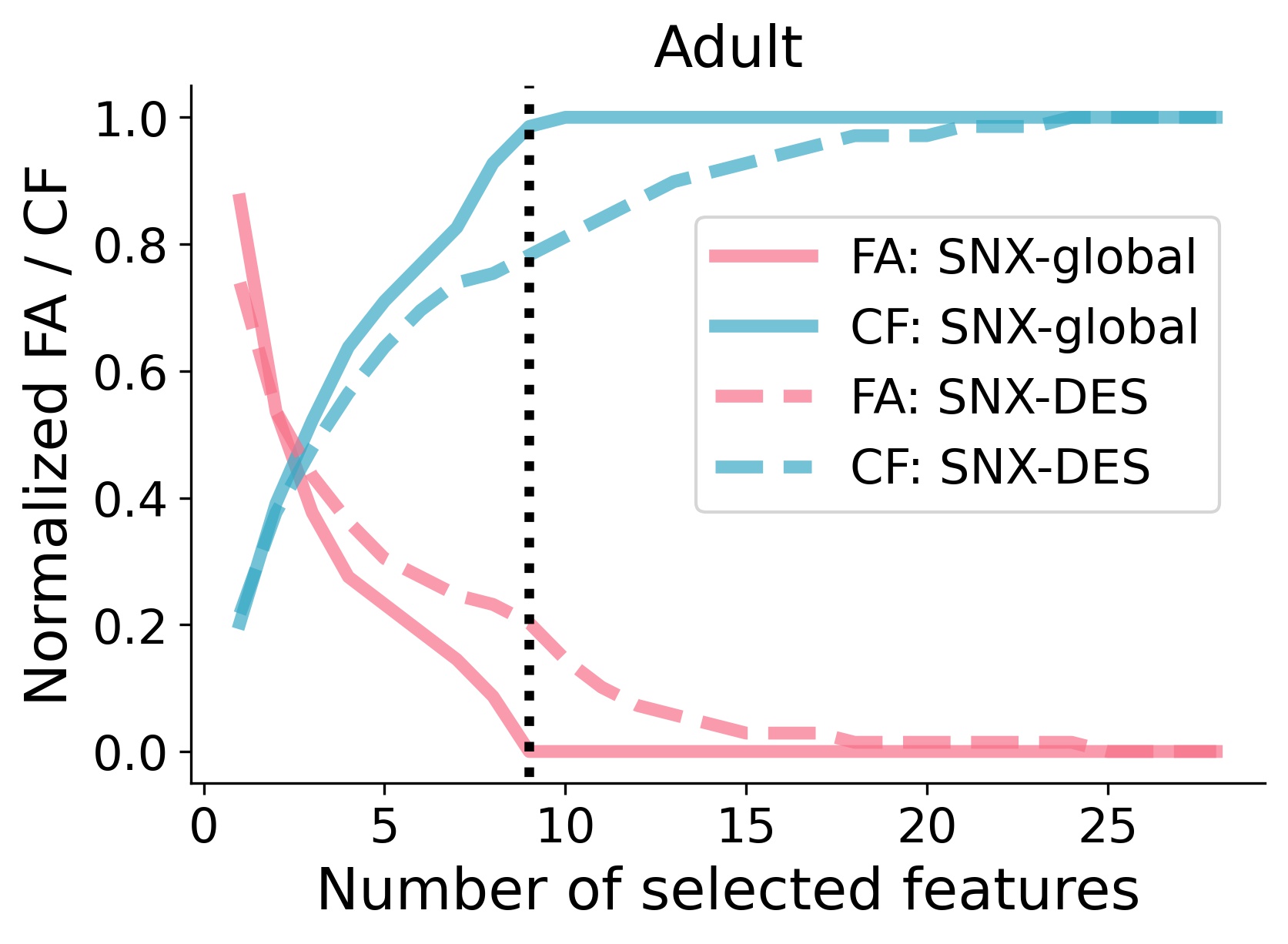}
\end{minipage}
\hspace{.2in}
\begin{minipage}{.2\textwidth}
    \includegraphics[width=\textwidth]{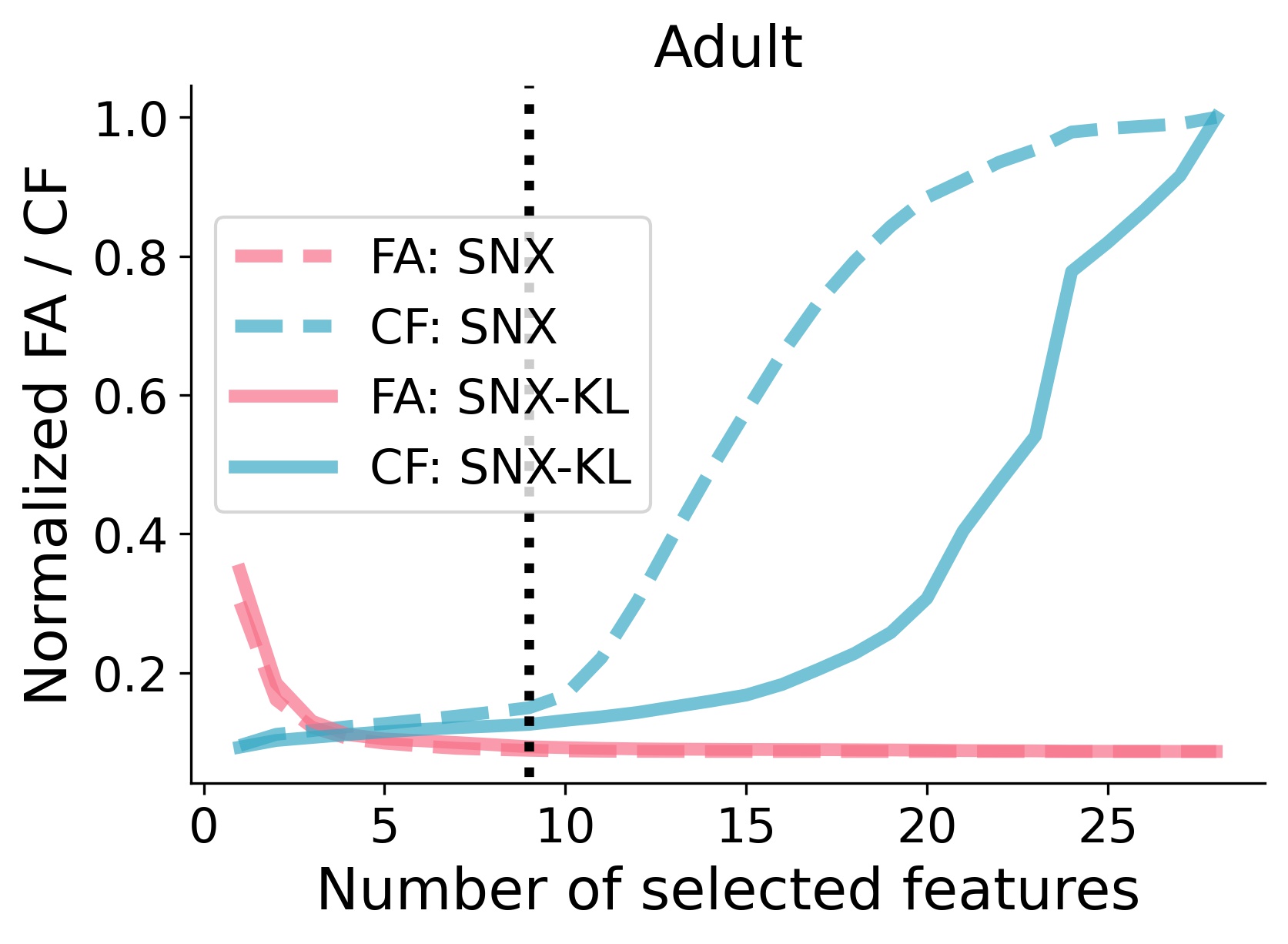}
\end{minipage}\\
\begin{minipage}{.2\textwidth}
    \includegraphics[width=\textwidth]{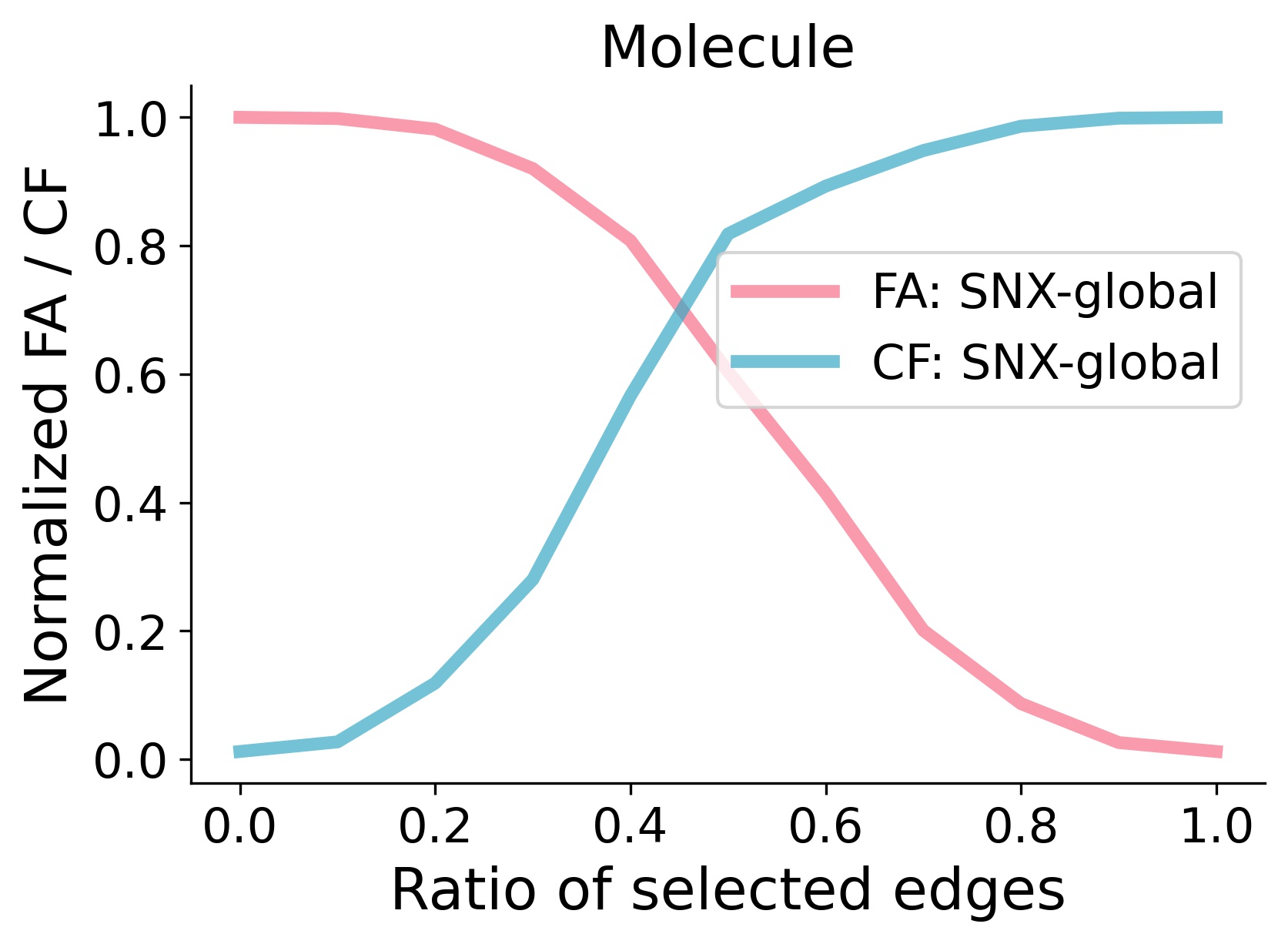}
\end{minipage}
\hspace{.2in}
\begin{minipage}{.2\textwidth}
    \includegraphics[width=\textwidth]{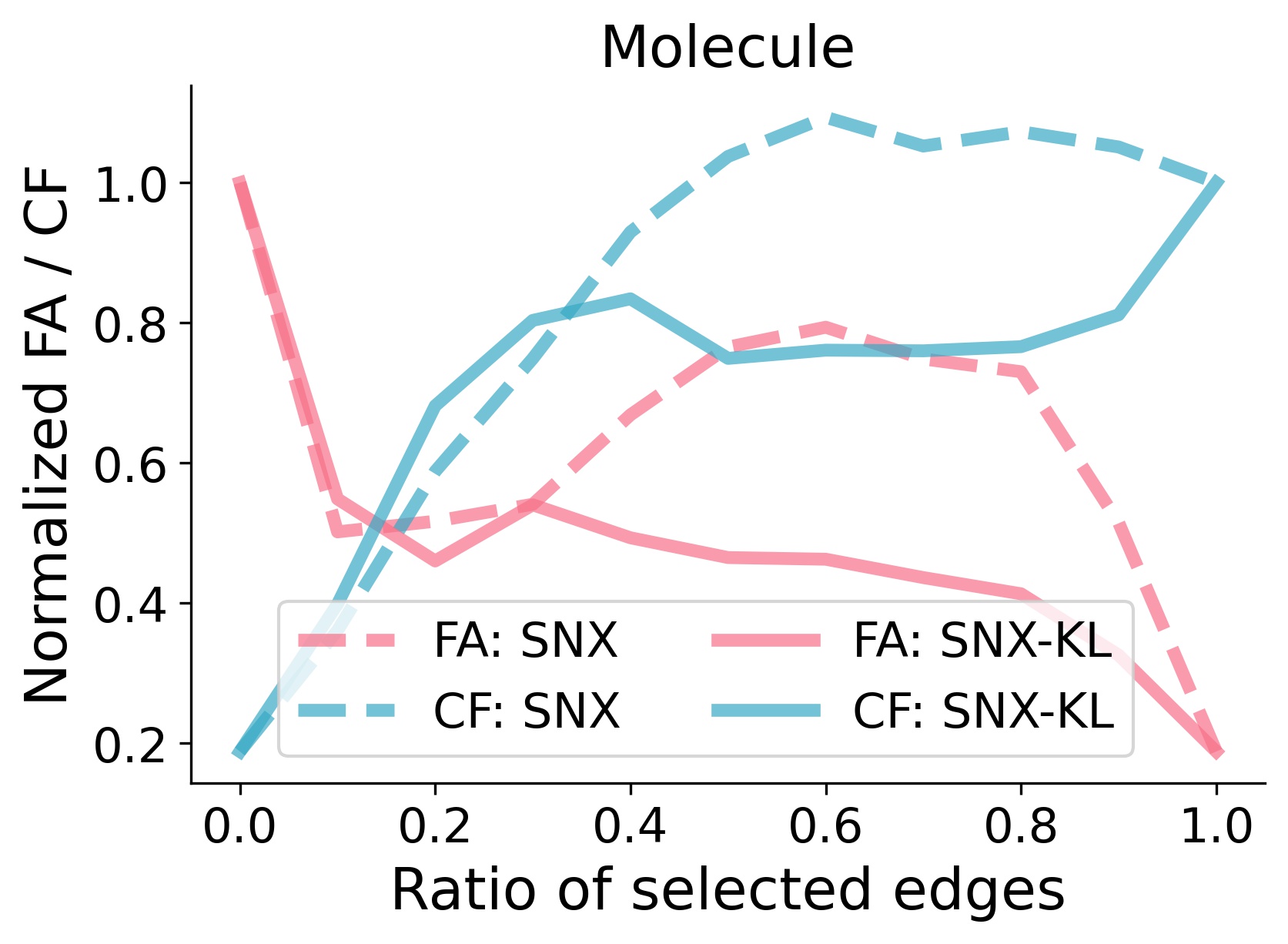}
\end{minipage}\\
    \caption{\small 
    Sensitivity of global (left) and local (right) masks w.r.t. faithfulness (FA) and counterfactual (CF). 
    %
    %
    }
    \label{fig:sensitivity}
\end{figure}
In Fig.~\ref{fig:sensitivity},
we plot the quality of the global masks (left column) and local masks (right column) on a tabular dataset (top row) and a graph dataset (bottom row).
Regarding global masks, on the Adult dataset, SNX-global outperforms DES and remains good even when smaller than 10 minor features are selected.
On the Molecule dataset, 
a higher percentage of edges need to be selected to preserve the connection patterns that are critical to GNN.
Regarding local masks, a much smaller number of minor features are needed to achieve good faithfulness (red curves for FA go down more rapidly), though more are needed to cover \textit{all} salient minor features (green curves for CF going up more slowly).
On the graph dataset, the stage 2 optimization can locate a smaller percentage of edges to preserve SN predictions,
and
a higher percentage of edges are needed to include all important edges, 
as seen in the CF losses.
Similar trends are found in other datasets.
Comparing FA of the global and the local masks in Adult,
we found that the global mask selects more features (9) than local one (about 5) to reach the elbow point.
The reason is that global mask needs to figure out \textit{all} non-zero features but local one only picks the non-zero features indicating the (dis)similarity of two feature vectors.
Similar results can be found in Molecule.

\subsection{Qualitative evaluation}
\label{sec:quali}

\begin{table}[t]
    \footnotesize
    \centering
    \caption{\small Visualization of local masks extracted by SM, SNX-UC, and SNX when comparing a query with two different references.
    }
    \label{tab:simplified_case_study}
    \begin{tabular}{c | c | c c c c c c c c c c}
        \toprule
        \multicolumn{2}{c}{\textbf{Adult}} 
        & \rotatebox[origin=c]{90}{Work} 
        & \rotatebox[origin=c]{90}{Race} 
        & \rotatebox[origin=c]{90}{Edu.}
        & \rotatebox[origin=c]{90}{Age} 
        & \rotatebox[origin=c]{90}{Hrs/Wk} 
        & \rotatebox[origin=c]{90}{Marriage} 
        & \rotatebox[origin=c]{90}{Occup.} 
        & \rotatebox[origin=c]{90}{Relation} 
        & \rotatebox[origin=c]{90}{Sex} 
        & \rotatebox[origin=c]{90}{Label} \\
        \midrule
        \multirow{3}{*}{Ref. 1} 
        & SM
        & \cellcolor{red!75}
        & \cellcolor{red!15} 
        & \cellcolor{red!15} 
        & \cellcolor{red!75}
        & \cellcolor{red!45}
        & \cellcolor{red!15}
        & \cellcolor{red!45}
        & \cellcolor{red!75}
        & \cellcolor{red!45}
        & 0 \\
        & SNX-UC
        & \cellcolor{red!75}
        & \cellcolor{red!15}
        & \cellcolor{red!15}
        & \cellcolor{red!45}
        & \cellcolor{red!75}
        & \cellcolor{red!45}
        & \cellcolor{red!45}
        & \cellcolor{red!75}
        & \cellcolor{red!15}
        & 0 \\
        & SNX
        & \cellcolor{red!75}
        & \cellcolor{red!15}
        & \cellcolor{red!45}
        & \cellcolor{red!15}
        & \cellcolor{red!75}
        & \cellcolor{red!45}
        & \cellcolor{red!75}
        & \cellcolor{red!15}
        & \cellcolor{red!15}
        & 0 \\
        \midrule
        Query 
        & SNX-global
        & \cellcolor{blue!75} 
        & \cellcolor{blue!15} 
        & \cellcolor{blue!75} 
        & \cellcolor{blue!15} 
        & \cellcolor{blue!75} 
        & \cellcolor{blue!45} 
        & \cellcolor{blue!75} 
        & \cellcolor{blue!45} 
        & \cellcolor{blue!15} 
        & 0 \\
        \midrule
        \multirow{3}{*}{Ref. 2} 
        & SNX
        & \cellcolor{red!75} 
        & \cellcolor{red!15} 
        & \cellcolor{red!15} 
        & \cellcolor{red!15} 
        & \cellcolor{red!75} 
        & \cellcolor{red!45} 
        & \cellcolor{red!75} 
        & \cellcolor{red!45} 
        & \cellcolor{red!15} 
        & 1 \\
        & SNX-UC
        & \cellcolor{red!75} 
        & \cellcolor{red!15} 
        & \cellcolor{red!15} 
        & \cellcolor{red!45} 
        & \cellcolor{red!45} 
        & \cellcolor{red!75} 
        & \cellcolor{red!75} 
        & \cellcolor{red!45} 
        & \cellcolor{red!15} 
        & 1 \\
        & SM
        & \cellcolor{red!45} 
        & \cellcolor{red!75} 
        & \cellcolor{red!15} 
        & \cellcolor{red!45} 
        & \cellcolor{red!75} 
        & \cellcolor{red!15} 
        & \cellcolor{red!75} 
        & \cellcolor{red!45} 
        & \cellcolor{red!15} 
        & 1 \\
        \bottomrule
    \end{tabular}
\end{table}

\noindent\textbf{Adult Dataset.}
Table \ref{tab:simplified_case_study} shows a case study from the Adult dataset.
The global invariant mask $\mathbf{N}^s$ over major features $\mathbf{z}^s$ (Query) is highlighted in blue.
Similarly, local masks $\mathbf{n}$ over major feature vectors extracted by SM, SNX-UC, and SNX when comparing query with Ref.1 and Ref.2 are colored red.
Higher color saturation indicates more saliency.
It is clear that SNX closely follows the constraints set by the global mask  $\mathbf{N}^s$:
SNX always places the same or lower importance on a feature than that place by SNX-global.
On the contrary,
SNX-UC and SM do not consider constraints and both highlight some features,
such as  \texttt{Race} and \texttt{Age},
that are not considered important by SNX-global.
When fairness is a concern,
not selecting sensitive features such as \texttt{Race}, \texttt{Age}, and \texttt{Sex} can be a constraint built in the global mask $\mathbf{N}^s$.
Without such domain requirements as labeled data, it is encouraging to see the self-supervised learning can avoid selecting these sensitive features.
We note that whether the query is compared with references in the same or different classes, 
SNX conforms to the global invariant.

\begin{figure*}
    \centering
\begin{minipage}{.2\textwidth}
    \includegraphics[width=\textwidth]{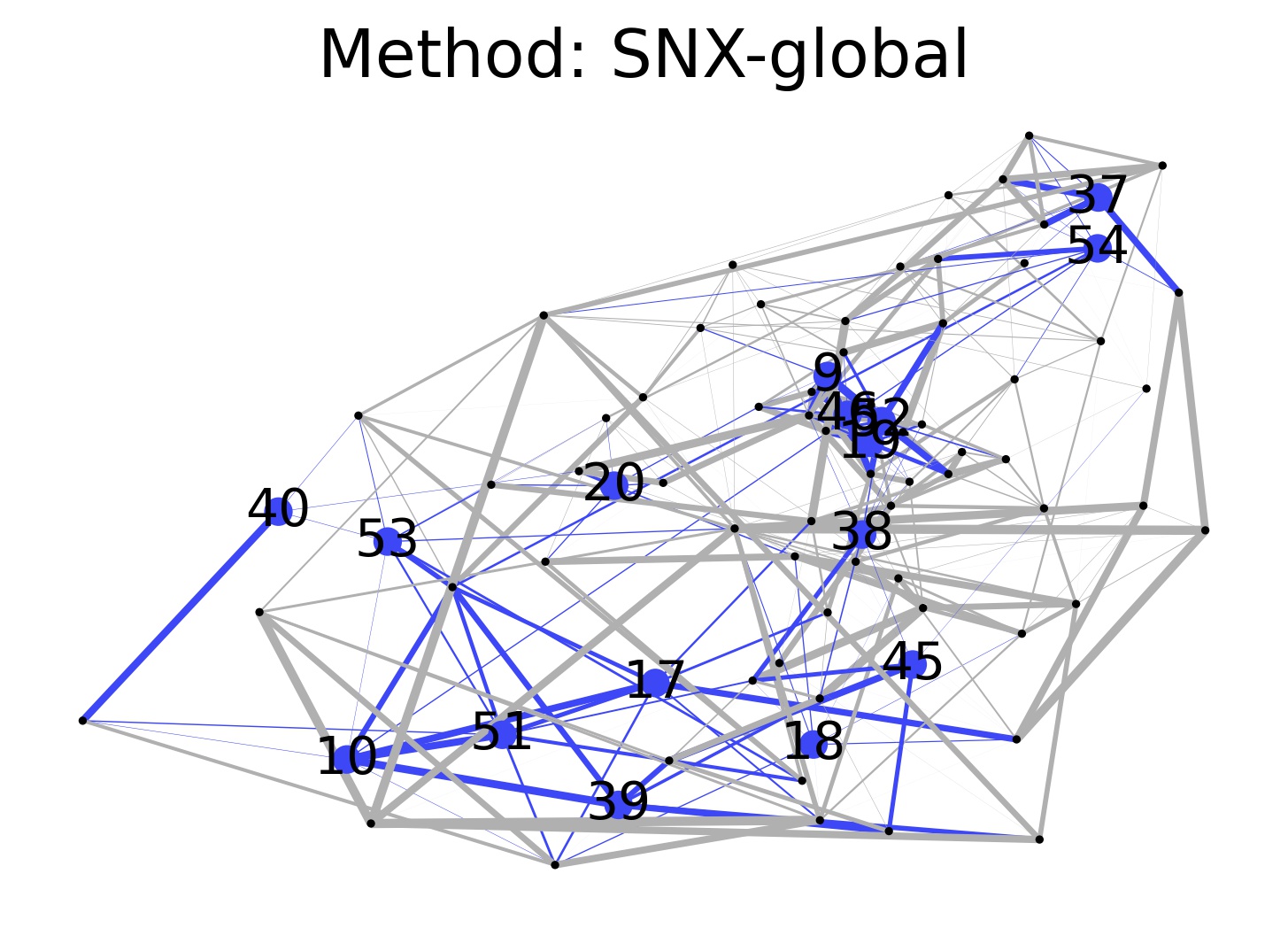}
\end{minipage}%
\begin{minipage}{.2\textwidth}
    \includegraphics[width=\textwidth]{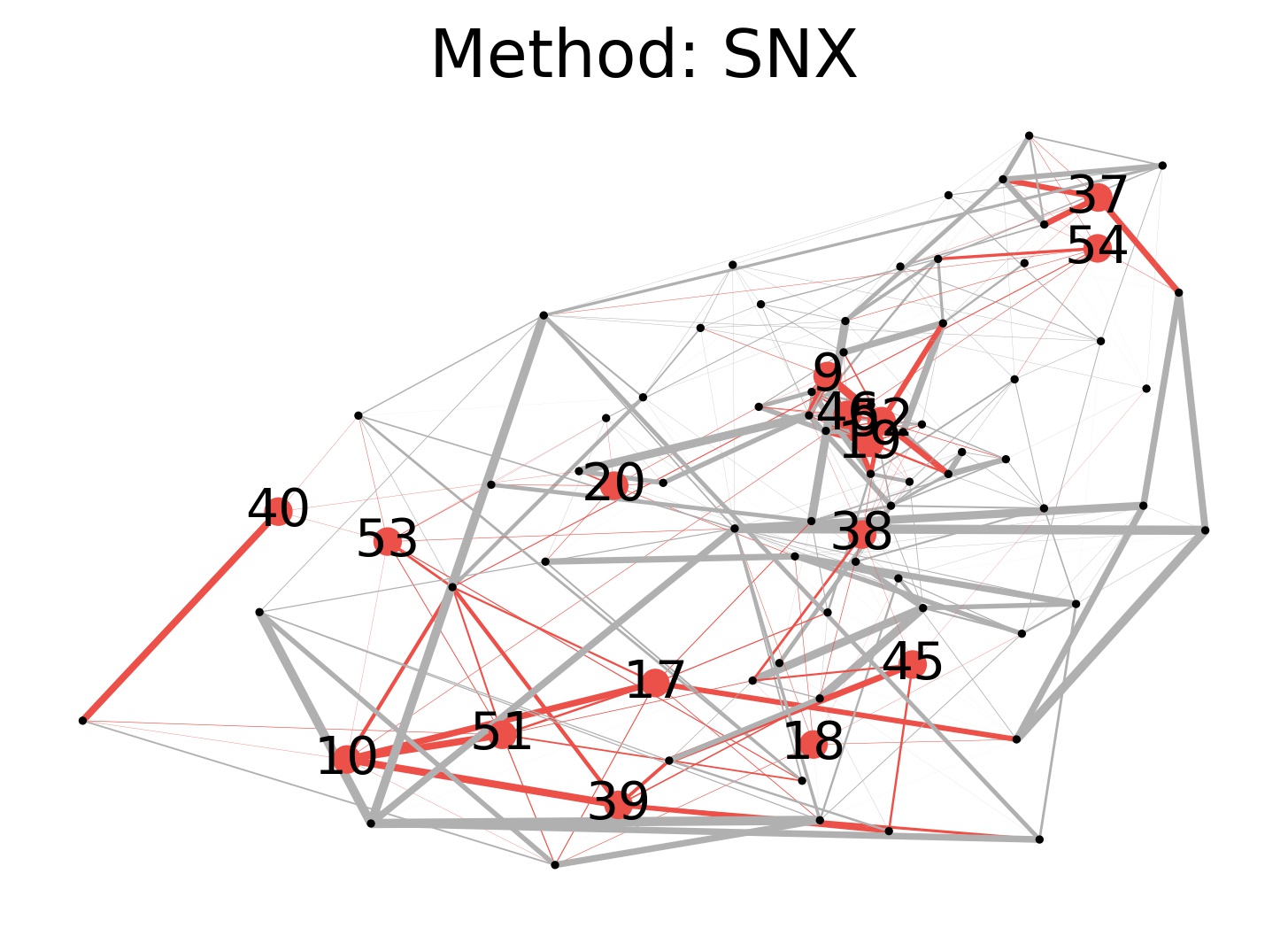}
\end{minipage}%
\begin{minipage}{.2\textwidth}
    \includegraphics[width=\textwidth]{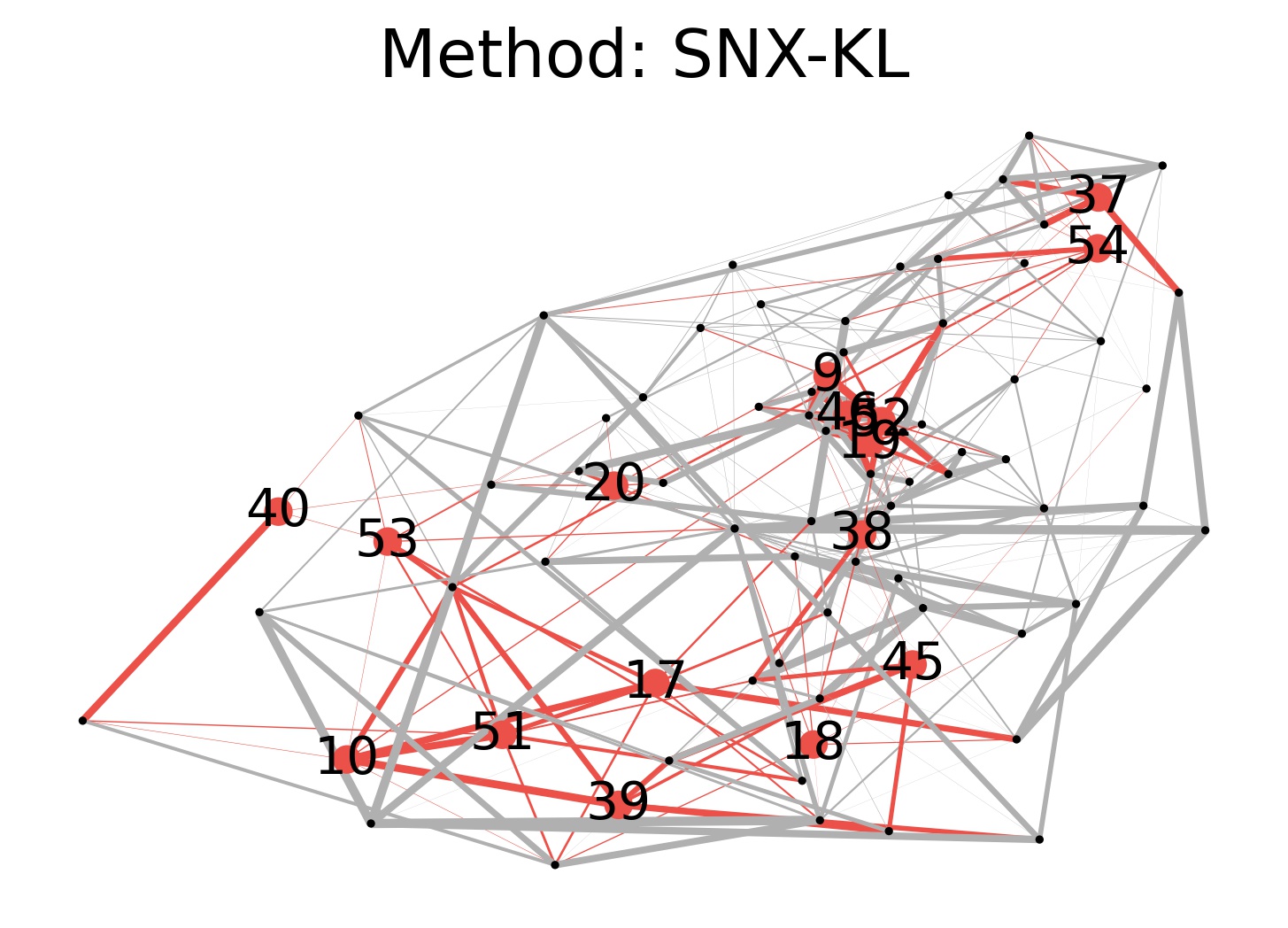}
\end{minipage}%
\begin{minipage}{.2\textwidth}
    \includegraphics[width=\textwidth]{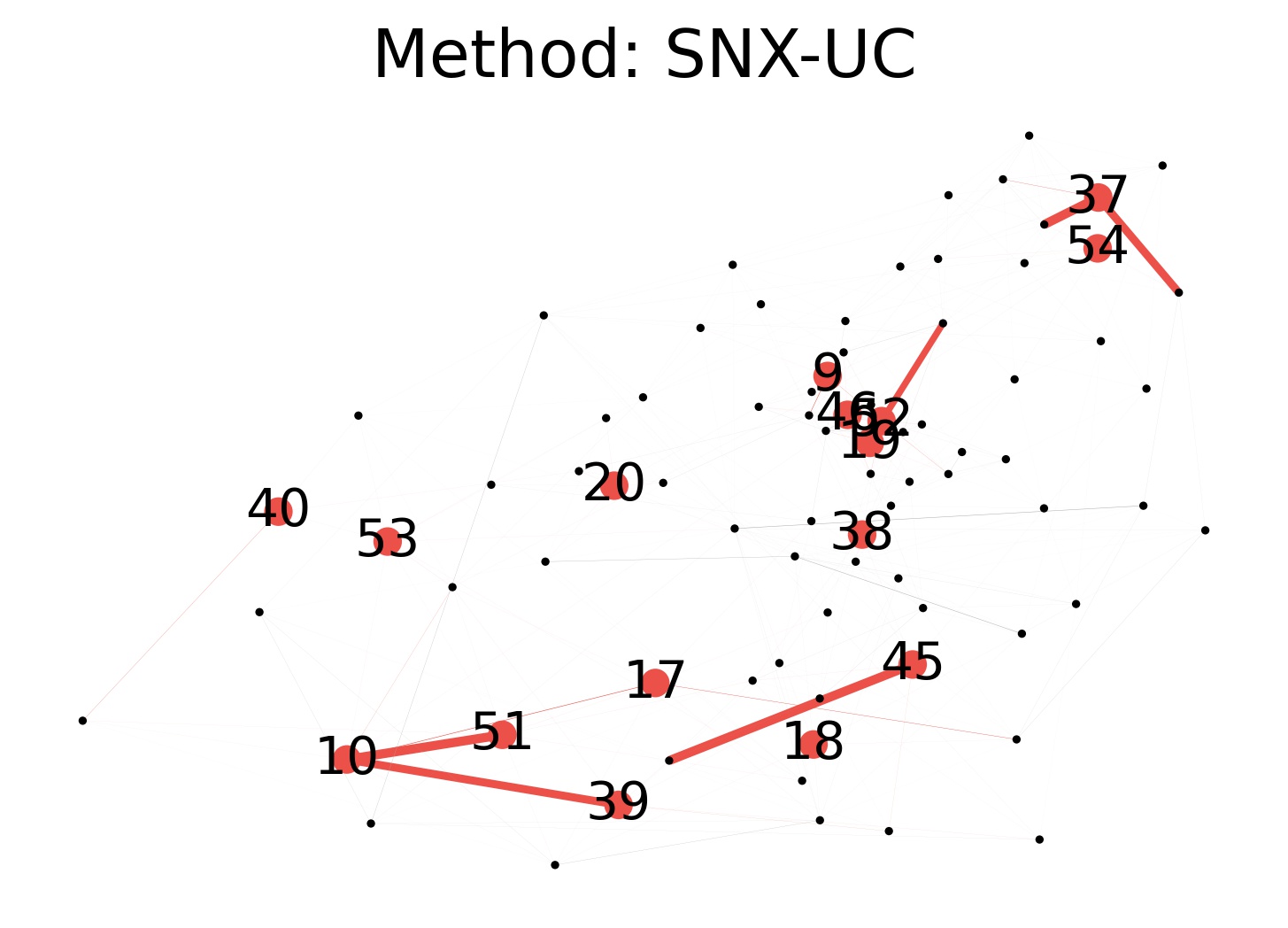}
\end{minipage}%
\begin{minipage}{.2\textwidth}
    \includegraphics[width=\textwidth]{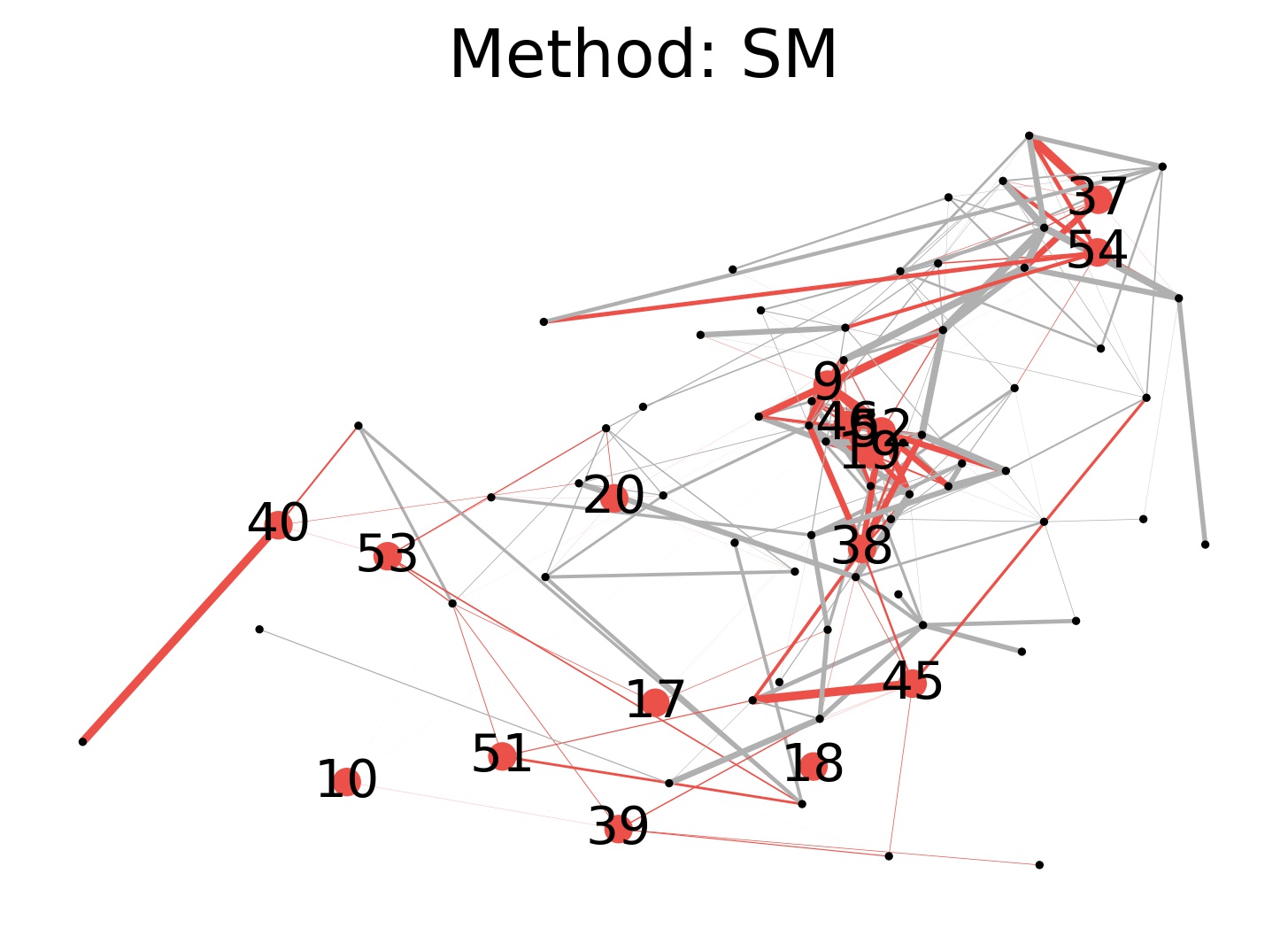}
\end{minipage}\\
    \caption{\small 
    Explaining brain network matching in the BP dataset.
    The numbered nodes and represent the brain ROIs.
    Blue highlights globally salient edges and red highlights locally salient edges.
    Edge thickness indicates the importance of edges in different masks.
    SNX-global found important edges linking ROIs (first subfigure),
    and both SNX-KL and SNX found local masks conforming to the global mask.
    }
    \label{fig:bp_case_study}
\end{figure*}

\begin{figure}[t]
    \centering
\begin{minipage}{.23\textwidth}
    \includegraphics[width=\textwidth]{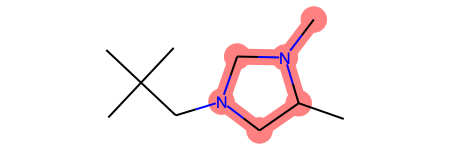}
\end{minipage}%
\begin{minipage}{.23\textwidth}
    \includegraphics[width=\textwidth]{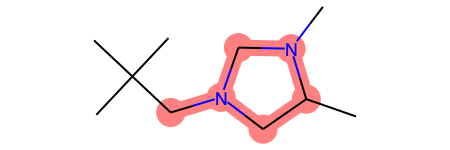}
\end{minipage} \\
\begin{minipage}{.23\textwidth}
    \includegraphics[width=\textwidth]{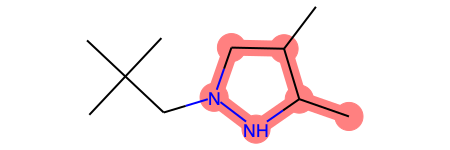}
\end{minipage} %
\begin{minipage}{.23\textwidth}
    \includegraphics[width=\textwidth]{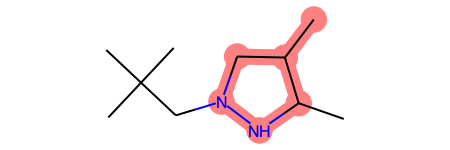}
\end{minipage}%
    \caption{\small Case study for 
    molecules.
    Each highlighted subgraph appears in two promising molecules and never in the useless molecules.
    }
    \label{fig:case-study-molecule}
\end{figure}

\noindent\textbf{Brain networks.}
Fig.~\ref{fig:bp_case_study} shows masks learned by SNX-global and four local methods on the BP dataset, respectively.
The numbered colored nodes are brain regions of interest (ROIs) related to the human dorsal and ventral systems, 
and edges connected to these nodes are colored accordingly. 
These ROIs,
such as ventromedial prefrontal cortex (\#39), 
dorsolateral prefrontal cortex (\#17), 
superior parietal lobule (\#10), 
and anterior cingulate cortex (\#51),
could be highly affected by bipolar disorder according to neuroscience studies \cite{chen2011quantitative,zovetti2020default}.
As a result, 
SNX-global well captures many edges adjoined to these ROIs,
particularly the interconnections among the aforementioned four ROIs (\#39, \#17, \#10 and \#51).
Also, SNX and SNX-KL found local masks more conformal to global mask than the baselines SNX-UC and SM.


\noindent\textbf{Chemical molecules.}
We extract all the 38 pairs of connected cliques that appear in the explaining subgraphs. After consulting with one of the authors who is a chemical engineer, we found the salient pairs of cliques that appear only in the positive or negative molecules. We display these fragments in Fig.~\ref{fig:case-study-molecule}, 
along with their frequencies in the entire dataset. 
For example, heteroatoms (i.e., O and N atoms) that are a part of a ring are often found in positive molecules. 
Further, interestingly, consecutive heteroatoms within rings (NN and OO) were also found to be only in positive molecules.
The algorithm did identify fragments, such as ``CC-C1COCN'', that appear only in the negative class, but belong to motifs that might be considered favorable. We posit that the reason for this is that the junction tree cannot accurately specify how two cliques are connected (i.e., to which atom of the ring "C1COCN1" is the fragment "CC" connected to), 
and that the dataset used to train the GNN may not be sufficiently large and diverse due to its construction~\cite{paragian2020computational}.

\subsection{Reproducibility checklist}
\label{sec:repro-list}
\noindent \textbf{Pairs generation.}
For all six datasets, 
in both training portion and test portion,
we pair each sample with other 4 randomly selected samples in the same portion,
where 2 are from the same class and the other 2 from the different class.
\noindent \textbf{Hypreparameters settings.}
For all experiments, 
$\eta_1=10^{-1},\eta_2=10^{-3}$ in Eq. (\ref{eq:variable_update})
and $\beta=1$ in SNX-KL.
On the tabular datasets,
$\gamma=10^{-3}$ in both Eq. (\ref{eq:obj_gm}) 
and Eq. (\ref{eq:con_lm_obj}).
To extract global masks,
learning rate is $10^{-1}$, 
and $MaxIter=50$;
as for local masks,
$PreIter=50, MaxIter=100$.
On graph datasets, 
$\gamma$ in Eq. (\ref{eq:obj_gm}) and Eq. (\ref{eq:con_lm_obj}) is $10^{-1}$ for Molecule and $10^{-4}$ for BP.
To extract global masks,
learning rate is $10^{-1}$, and $MaxIter=200$;
as for local masks,
$PreIter=0, MaxIter=400$.

\section{Related work}
\label{sec:related}
Explainability in machine learning can be attained by intrinsically transparent models~\cite{Lakkaraju2016,Lou2012} or simpler surrogate models~\cite{ribeiro2016should}.
Global explanations in prior work mean ``regardless of input data'', while we define global explanations differently for SN, meaning ``regardless of the reference instance.''
In~\cite{Rudin2019GloballyConsistentRS}, the authors 
use global consistency constraints to regularize local explanations, with different definitions of ``global consistency'' and target model architecture.
Explainability have been extensively studied for tabular data~\cite{ribeiro2016should}, images~\cite{selvaraju2017grad,dhurandhar2018explanations}, and texts~\cite{jain2019}, and more recently, on graph data~\cite{ying2019gnnexplainer,baldassarre2019explainability,luo2020parameterized,fabercontrastive,Yuan2020XGNNTM,Vu2020PGMExplainerPG,Lanciano2020ExplainableCO,yuan2020explainability}.

Robustness in explanations is gaining attention~\cite{Dombrowski2020TowardsRE,lakkaraju20a,Wang2020SmoothedGF}.
In~\cite{Dombrowski2020TowardsRE}, the goal is to train neural networks for image classification that has robust explanations with malicious data manipulations.
In~\cite{Wang2020SmoothedGF}, the vulnerability of explanations of architecture with a single input image is analyzed.
In the prior work, explanation robustness is analyzed local to a neighborhood of the single input $\mathbf{x}$,
while we characterize robustness with respect to varying references in SN.



Explaining why two instances are similar or different has been only sparsely researched
\cite{fabercontrastive, Lanciano2020ExplainableCO,plummer2019these,utkin2019explanation}.
\cite{fabercontrastive} extracts the most contrastive parts of the graphs to tell the similarity (or difference) among those in the same class (or different classes).
\cite{Lanciano2020ExplainableCO} extracts contrast subgraphs for discriminating two different groups of brain networks. 
In \cite{plummer2019these}, salient attributes are introduced to explain image similarity. 
\cite{utkin2019explanation} proposes to compare the explained example with a prototype in the embedding space of the SN and reconstruction is needed, while we learn masks on the input directly.

The convergence of GDA has been extensively studied in optimization, game theory, adversarial training, and security.
The closest work to our work is~\cite{Wachter2017}, where the authors used GDA to solve a constrained optimization problem to find counterfactual explanations for a classifier.
Rather, we apply the algorithm to solve a novel 
problem for SN explanation.
\section{Conclusions}
\label{sec:conclusions}
We address the robustness issues in explaining Siamese networks due to changing compared reference objects.
We formulated the problem of global invariance when explaining the results of comparing two objects,
where the invariance is self-learned from unlabeled data.
The optimization is done using SGD and GDA that are provably and empirically convergent to high-quality solutions.
Case studies on a tabular dataset and human brain networks demonstrate the meaningfulness of the learned invariant and conformity when explaining local comparisons.
\bibliographystyle{plain}
\bibliography{paper}

\end{document}